\documentclass[10pt,twocolumn,letterpaper]{article}

\usepackage[pagenumbers]{cvpr} %

\usepackage[dvipsnames]{xcolor}

\usepackage{mathtools}

\newcommand{\algabbr}{GARField}
\newcommand{\grouping}{scale-conditioned affinity} %
\newcommand{\algfull}{Group Anything with Radiance Fields}

\usepackage{array,multirow,multicol}

\definecolor{cvprblue}{rgb}{0.21,0.49,0.74}
\usepackage[pagebackref,breaklinks,colorlinks,citecolor=cvprblue]{hyperref}

\def\paperID{8615} %
\def\confName{CVPR}
\def\confYear{2024}

\title{
\algabbr{}: \algfull{}
}
\author{Chung Min Kim$^{*1}$\; Mingxuan Wu$^{*1}$\; Justin Kerr$^{*1}$\; Ken Goldberg$^1$\; \\Matthew Tancik$^2$\;  Angjoo Kanazawa$^1$ \\
$*$ Denotes equal contribution
\\
$^1$UC Berkeley \hspace{1cm} \hspace{1cm} $^2$ Luma AI
}

\begin{document}

\twocolumn[
{
\renewcommand\twocolumn[1][]{#1}%
\maketitle
\begin{center}
\vspace{-2em}
\includegraphics[width=\textwidth]{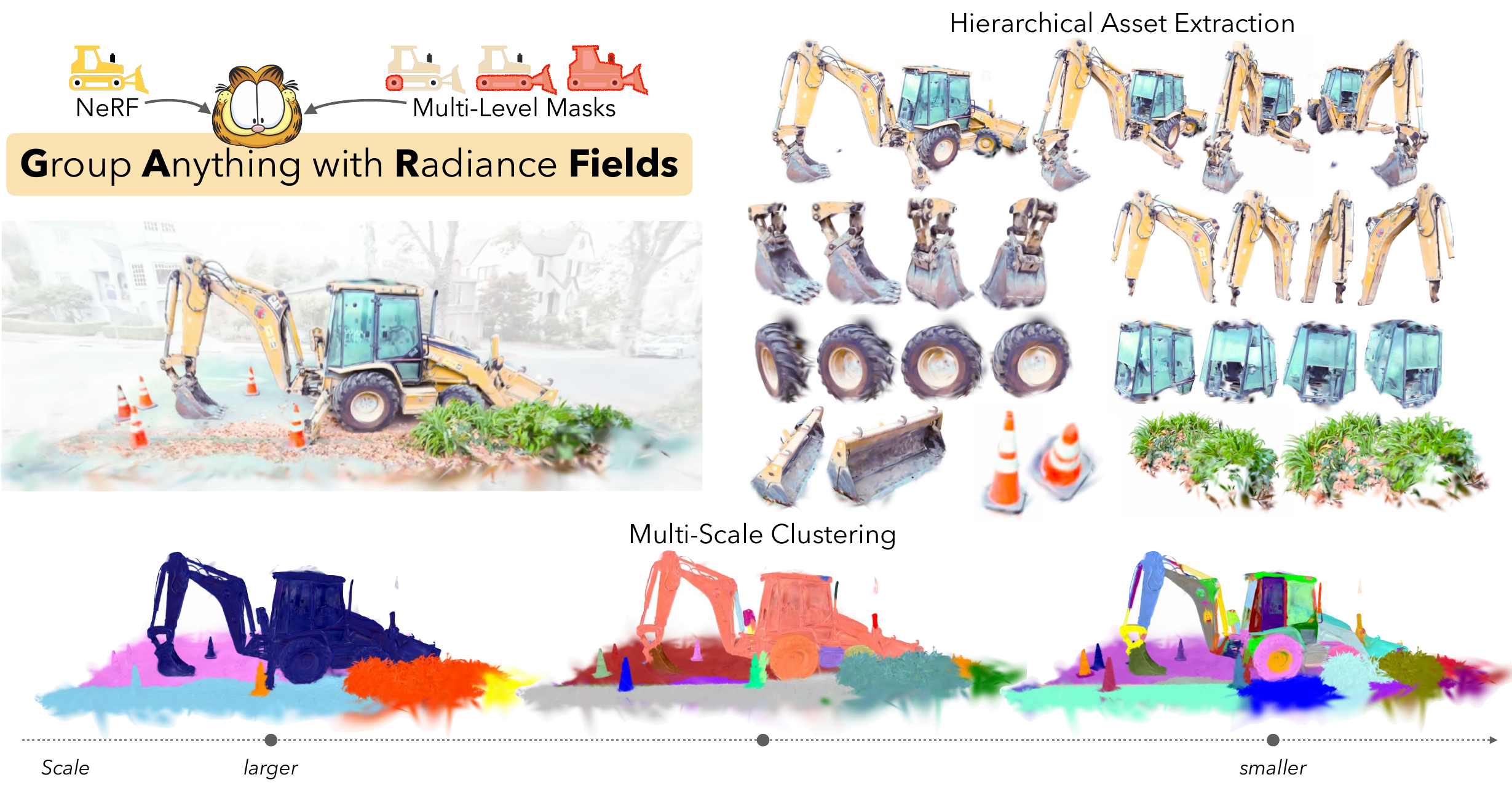}
\captionof{figure}{\textbf{\algfull{} (\algabbr{})} We present \algabbr{}, which distills multi-level groups represented as masks into NeRF to create a scale-conditioned 3D affinity field (top left). Once trained, this affinity field can be clustered at a variety of scales to decompose the scene at different levels of granularity, like breaking apart the excavator into its subparts (bottom). 3D assets can be extracted from this hierarchy by extracting every group in the scene automatically or via user clicks, as visualized here (top right).
}
\label{fig:teaser}
\end{center}
}]

\begin{abstract}
\vspace{-1em}Grouping is inherently ambiguous due to the multiple levels of granularity in which one can decompose a scene---should the wheels of an excavator be considered separate or part of the whole?
We present Group Anything with Radiance Fields (GARField), an approach for decomposing 3D scenes into a hierarchy of semantically meaningful groups from posed image inputs.
To do this we embrace group ambiguity through physical scale: by optimizing a scale-conditioned 3D affinity feature field, 
a point in the world can belong to different groups of different sizes. We optimize this field from a set of 2D masks provided by Segment Anything (SAM) in a way that respects coarse-to-fine hierarchy, using scale to consistently fuse conflicting masks from different viewpoints.
From this field we can derive a hierarchy of possible groupings via automatic tree construction or user interaction. We evaluate \algabbr{} on a variety of in-the-wild scenes and find it effectively extracts groups at many levels: clusters of objects, objects, and various subparts.
\algabbr{} inherently represents multi-view consistent groupings and produces higher fidelity groups than the input SAM masks. 
\algabbr{}'s hierarchical grouping could have exciting downstream applications such as 3D asset extraction or dynamic scene understanding. See the project website at \url{https://www.garfield.studio/}

\end{abstract}
\vspace{-1em}    
\section{Introduction}
\label{sec:intro}
Consider the scene in Figure~\ref{fig:teaser}. Though recent technologies like NeRFs~\cite{mildenhall2020nerf} can recover photorealistic 3D reconstructions of this scene, the world is modeled as a single volume with no structural meaning.
As humans, not only can we reconstruct the scene, but we also have the ability to \textit{group} it at multiple levels of granularity --- at the highest level, we see the parts of the scene \ie the excavator, bushes, and the sidewalk, but we are also able to decompose the excavator into its parts such as its wheels, crane, and the cabin. 
This ability to perceive the scene at multiple levels of groupings is a key component of our scene understanding, enabling us to interact with the 3D world by understanding what belongs together.
However, these different levels of granularity introduce ambiguity in groups, making it a challenge to represent them in a coherent 3D representation.
While there are multiple ways to break this ambiguity, we focus on the physical scale of entities as a cue to consolidate groups into a \textit{hierarchy}. 

In this work we introduce \algfull{} (\algabbr{}), an approach that, given posed images, reconstructs a 3D scene along with a scale-conditioned affinity field that enables decomposing the scene into a hierarchy of groups. 
For example, \algabbr{} can extract both the entire excavator (Fig.~\ref{fig:teaser} Top Right) as well as its subparts (Bottom Right). This dense hierarchical 3D grouping enables applications such as 3D asset extraction and interactive segmentation. 

\begin{figure}
    \includegraphics[width=\linewidth]{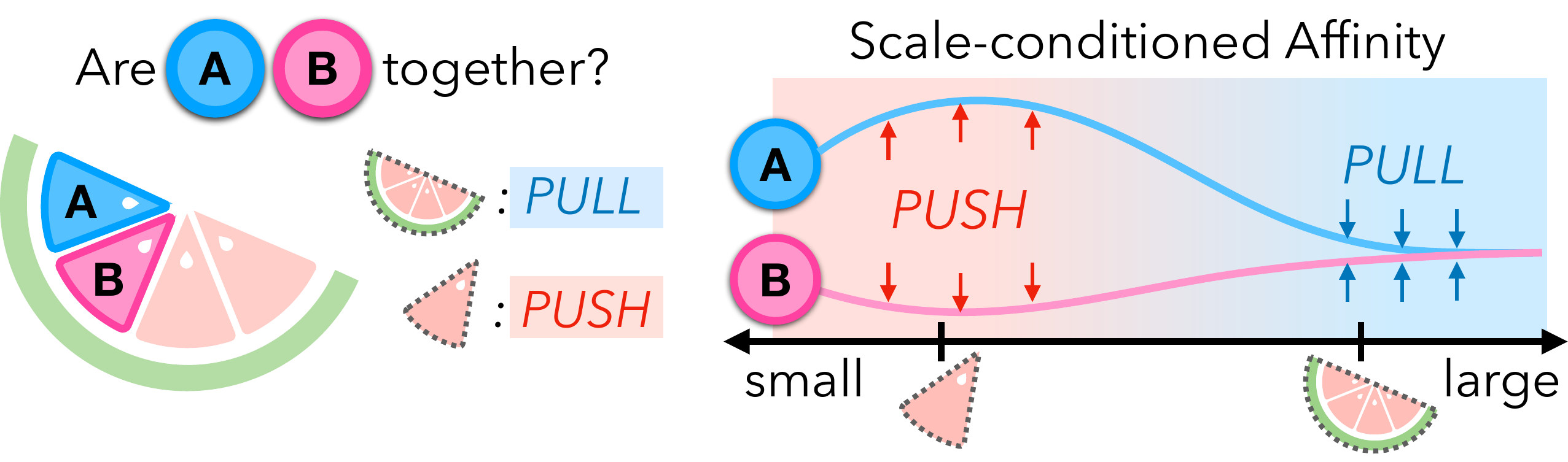}
    \caption{\textbf{Importance of Scale When Grouping}
A single point may belong to multiple groups. \algabbr{} uses \textit{scale-conditioning} to reconcile these conflicting signals into one affinity field. \vspace{-1em}}
 \label{fig:multiple-salient}\end{figure}

\algabbr{} distills a set of 2D segmentation masks into a 3D volumetric scale-conditioned affinity field. Because grouping is an ambiguous task, these 2D labels can be overlapping or conflicting. 
These inconsistencies pose a challenge for distilling masks into consistent 3D groups. We overcome this issue by leveraging a \textit{scale-conditioned} feature field. 
Specifically \algabbr{} optimizes a dense 3D feature field which is supervised such that feature distance reflects points' affinity. The scale conditioning enables two points to have higher affinity at a large scale but low affinity at a smaller scale (\ie wedges of the same watermelon), as illustrated in Figure~\ref{fig:multiple-salient}.

Though in principle \algabbr{} can distill any source of 2D masks, we derive mask candidates from Segment Anything Model (SAM)~\cite{kirillov2023segment} because they align well with what humans consider as reasonable groups. We process input images with SAM to obtain a set of candidate segmentation masks. For each mask, we compute a physical scale based on the scene geometry. To train \algabbr{}, we distill candidate 2D masks with a contrastive loss based on mask membership, leveraging 3D scale to resolve inconsistencies between views or mask candidates.

A well-behaved affinity field has: 1) \textit{transitivity} , which means if two points are mutually grouped with a third, they should themselves be grouped together, and 2) \textit{containment}, which means if two points are grouped at a small scale, they should be grouped together at higher scales.
\algabbr{}'s use of contrastive loss in addition to a containment auxiliary loss encourages both of these properties.

With the optimized scale-conditioned affinity field, \algabbr{} extracts a 3D scene hierarchy via recursively clustering them at descending scales until no more clusters emerge. By construction, this recursive clustering ensures that generated groups are subparts of the prior cluster in a coarse-to-fine manner.
We evaluate \algabbr{} on a variety of real scenes with annotated hierarchical groupings, testing its ability to capture object hierarchy, and its consistency across different views. By leveraging multiple views, \algabbr{} is able to produce detailed groupings, often improving upon the quality of input 2D segmentation masks. 
Moreover, these groups are 3D consistent by design, while 2D baselines do not guarantee view consistency.
We show downstream applications of \algabbr{} for 
hierarchical 3D asset extraction and click-based interactive segmentation. Given \algabbr{}'s scene decomposition capabilities, we're hopeful for its potential in other downstream applications like enabling robots to understand they can interact with or as a prior for dynamic reconstruction. 
Code and data will be released upon publication. Please see the supplemental video for more visualizations.

\section{Related Work}
\begin{figure*}
    \centering
    \includegraphics[width=\linewidth]{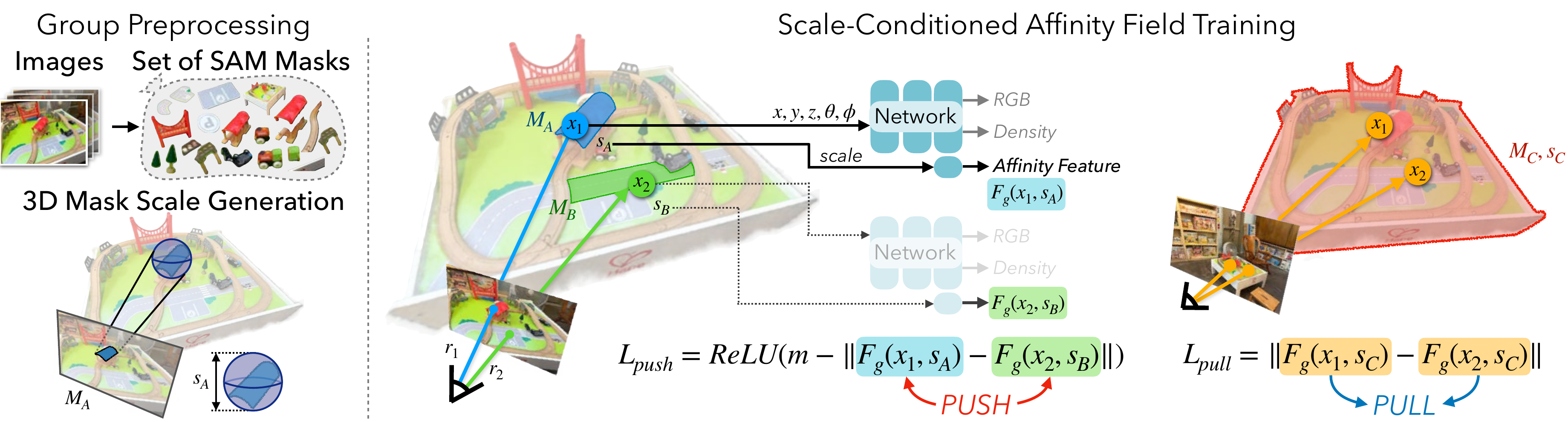}
    \caption{\textbf{\algabbr{} Method}: (Left) given an input image set, we extract a set of candidate groups by densely querying SAM, and assign each a physical scale by deprojecting depth from the NeRF. These scales are used to train a \textit{scale-conditioned affinity field} (Right). During training, pairs of sampled rays are pushed apart if they reside in different masks, and pulled together if they land in the same mask. Affinity is supervised only at the scale of each mask, which helps resolve conflicts between them.}
    \label{fig:method}
    \vspace{-1em}
\end{figure*}

\paragraph{Hierarchical Grouping} 
Multi-level grouping has long been studied in 2D images since the early days of foreground segmentation~\cite{ncuts}. Several methods build on this idea of spectral clustering for multi-level segmentation~\cite{cour2005} and more complex hierarchical scene parsing~\cite{UCMpaper,pont2016multiscale,socher2011parsing}. These approaches rely on extracting contours either via classic texture cues and create a hierarchy either via a top-down~\cite{yu2004segmentation} or bottom-up consolidation~\cite{UCMpaper}. More recent deep learning approaches use edges~\cite{xie2015holistically} computed at multiple scales to create the hierarchy, and Ke \etal~\cite{ke2022hsg} proposes a transformer based unsupervised hierarchical segmentation approach guided by the outputs of a classic hierarchical segmentation~\cite{UCMpaper}.

Many works circumvent the question of ambiguity in grouping by defining a set of categories within which instances are to be segmented, \ie panoptic segmentation~\cite{he2017mask,kirillov2019panoptic}. Recently, Segment Anything (SAM)~\cite{kirillov2023segment} off-loads this ambiguity into prompting, where at each pixel multiple segmentation masks can be proposed. However SAM does not recover a consistent set of hierarchical groups in the scene, which we enable by multi-scale 3D distillation. 

Hierarchical part decomposition has also been explored in 3D objects, either in a supervised~\cite{wang2011symmetry,li2017grass,mo2019structurenet}, or unsupervised manner~\cite{paschalidou2020learning}. Our approach distills information from a 2D model, and we consider full scenes while these approaches focus on 3D objects.

\noindent\textbf{Segmentation in NeRFs}
Existing approaches for segmentation in NeRFs typically distill segmentation masks into 3D either by using ground-truth semantic labels~\cite{semantic_nerf,panoptic_lifting}, matching instance masks~\cite{instancenerf}, or training 3D segmentation networks on NeRF~\cite{vora2021nesf}. However, these techniques do not consider hierarchical grouping, and are only interested in a flat hierarchy of objects or instances. \citet{ren2022neural} leverages human interaction in the form of image scribbles to segment objects with interaction. More recently, \citet{cen2023segment} try to recover a 3D consistent mask from SAM by tracking the 2D masks between neighboring views via user prompting. \citet{chen2023san} attempt this by distilling SAM encoder features into 3D and querying the decoder. 
In contrast with these approaches, our approach \algabbr{} 
does not require user input; it is able to obtain a hierarchical grouping of the scene automatically, 
and furthermore the recovered groups are view-consistent by definition.

\noindent\textbf{3D Feature Fields}
Distilling higher-dimensional features into a neural field, in tandem with a radiance field (view-dependent color and density), has been thoroughly explored. Methods like Semantic NeRF~\cite{semantic_nerf}, Distilled Feature Fields~\cite{kobayashi2022decomposing}, Neural Feature Fusion Fields~\cite{tschernezki22neural}, and Panoptic Lifting~\cite{panoptic_lifting} distill per-pixel 2D features into 3D by optimizing a 3D feature field to reconstruct the 2D features after volumetric rendering. These features can be either from pretrained vision models such as DINO or from semantic segmentation models.
LERF~\cite{lerf2023} extends this idea to a scale-conditioned feature field, enabling the training of feature fields from global image embeddings like CLIP~\cite{radford2021learning}. \algabbr{} similarly optimizes a scale-conditioned feature field in 3D; however, the purpose of the multi-scale features is to resolve ambiguity in grouping, instead of reconstructing an explicit 2D feature like CLIP. In addition LERF has no spatial grouping, a shortcoming \algabbr{} addresses.
\looseness=-1 The aforementioned methods are based on direct supervision from image features, while other methods such as NeRF-SOS~\cite{fan2022nerf} and Contrastive Lift~\cite{bhalgat2023contrastive} optimize an arbitrary feature field at a single scale using a contrastive loss between pairs of rays based on similarity. \algabbr{} uses this contrastive approach because it allows for defining pairwise relationships between points based on mask labels. However, we design a scale-conditioned contrastive loss, which allows for distilling conflicting masks into 3D. In addition, \algabbr{} does not require the slow-fast formulation of \citet{bhalgat2023contrastive} for stable training, perhaps enabled by scale-conditioned training.

\section{Method}

\subsection{2D Mask Generation}
\label{method:maskgen}

\algabbr{} takes as input a set of posed images and produces a hierarchical 3D grouping of the scene, along with a standard 3D volumetric radiance field and a scale-conditioned affinity field. To do this, we first pre-process input images with SAM to obtain mask candidates. Next, we optimize a volumetric radiance field along with the affinity field which takes in a single 3D location and a euclidean scale, and outputs a feature vector.
Affinity is obtained by comparing pairs of points' feature vectors.
After optimization, the resulting affinity field can be used to decompose a scene by recursively clustering the feature embeddings in 3D at descending scales in a coarse-to-fine manner, or for segmenting user specified queries. The overall pipeline is illustrated in Figure~\ref{fig:method}.

In order to train a \algabbr{}, we first mine 2D mask candidates from an image and then assign a 3D scale for each mask.
Specifically, we use SAM's automatic mask generator~\cite{kirillov2023segment}, which queries SAM in a grid of points and produces 3 candidate segmentation masks per query point. Then, it filters these masks by confidence and deduplicates nearly identical masks to produce a list of mask candidates of multiple sizes which can overlap or include each other. This process is done independently of viewpoint, producing masks which may not be consistent across views.
In this work we aim to generate a hierarchy of groupings based on objects' physical size. As such, we assign each 2D mask a physical 3D scale as in Fig.~\ref{fig:method}.
To do this we partially train a radiance field and render a depth image from each training camera pose. Next, for each mask we consider the 3D points within that mask and pick the scale based on the extent of the points' position distribution. This method ensures the 3D scale of masks resides in the same world-space, enabling scale-conditioned affinity.

\begin{figure}
        \includegraphics[width=\linewidth]{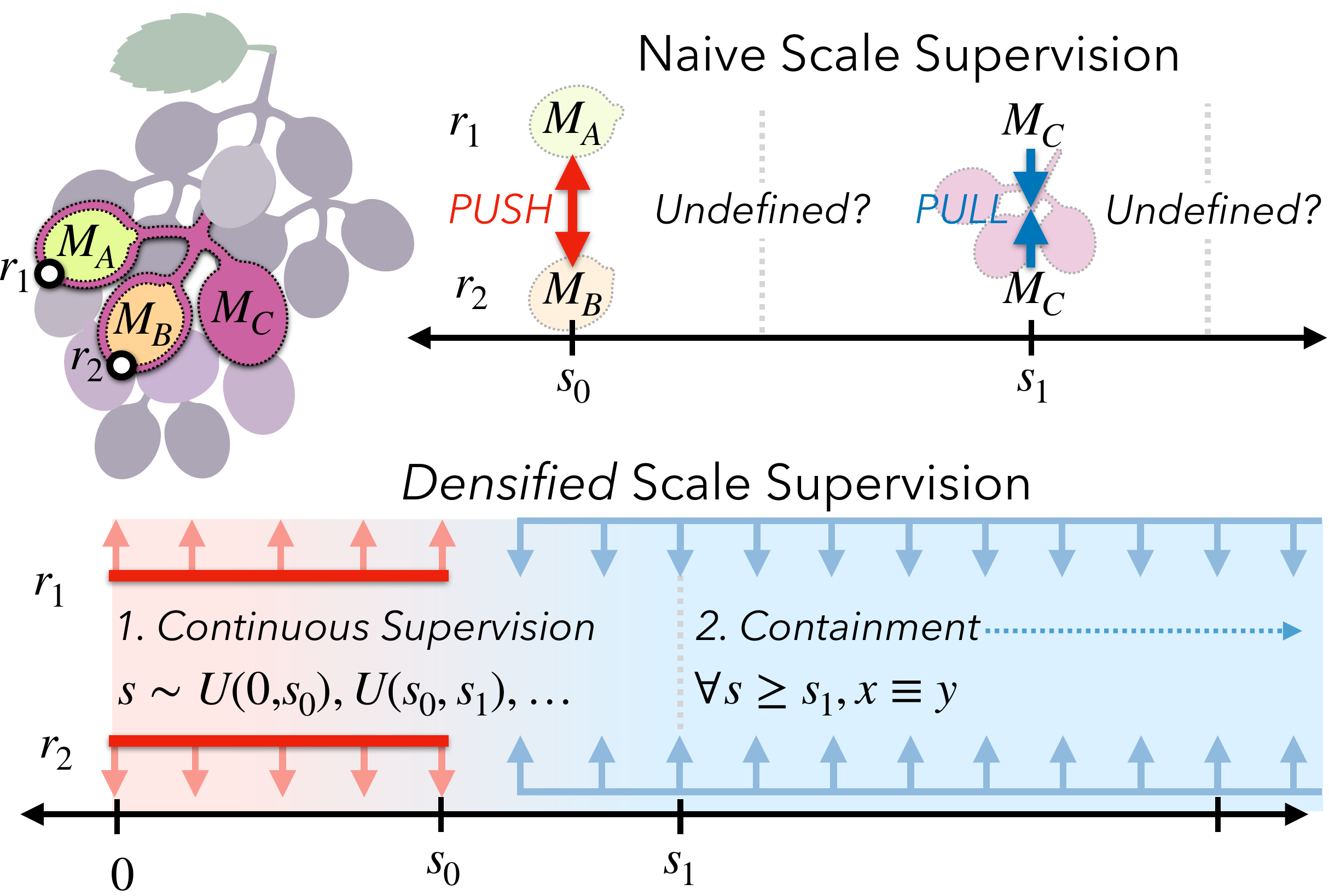}
        \caption{\textbf{Densified Scale Supervision}: Consider two grapes within a cluster. \textit{Naively} using scale for contrastive loss supervises affinities only at the grape and grape trio levels, leaving entire intervals unsupervised. 
     In \algabbr{}, we densify the supervision by 1) augmenting scale between mask euclidean scales and 2) imposing an auxiliary loss on containment of larger scales.} \label{fig:densification}
    \vspace{-1em}
\end{figure}

\subsection{Scale-Conditioned Affinity Field} 

Scale-conditioning is a key component of \algabbr{} which allows consolidating inconsistent 2D mask candidates: The same point may be grouped in several ways 
depending on the granularity of the groupings desired. Scale-conditioning alleviates this inconsistency because it resolves ambiguity over which group a query should belong to. Under scale-conditioning, conflicting masks of the same point no longer fight each other during training, but rather can coexist in the same scene at different affinity scales.

We define the \grouping{} field $F_{\text{g}}(x, s)\mapsto R^d$ over a 3D point $x$ and euclidean scale $s$, similar to the formulation in LERF~\cite{lerf2023}. Output features are constrained to a unit hyper-sphere, and the affinity between two points at a scale is defined by $A(x_1,x_2,s)=-||F_{\text{g}}(x_1, s) - F_{\text{g}}(x_2, s)||_2$. These features can be volumetrically rendered with a weighted average using the same rendering weights based on NeRF density to obtain a value on a per-ray basis.

\begin{figure}
    \centering
    \includegraphics[width=\linewidth]{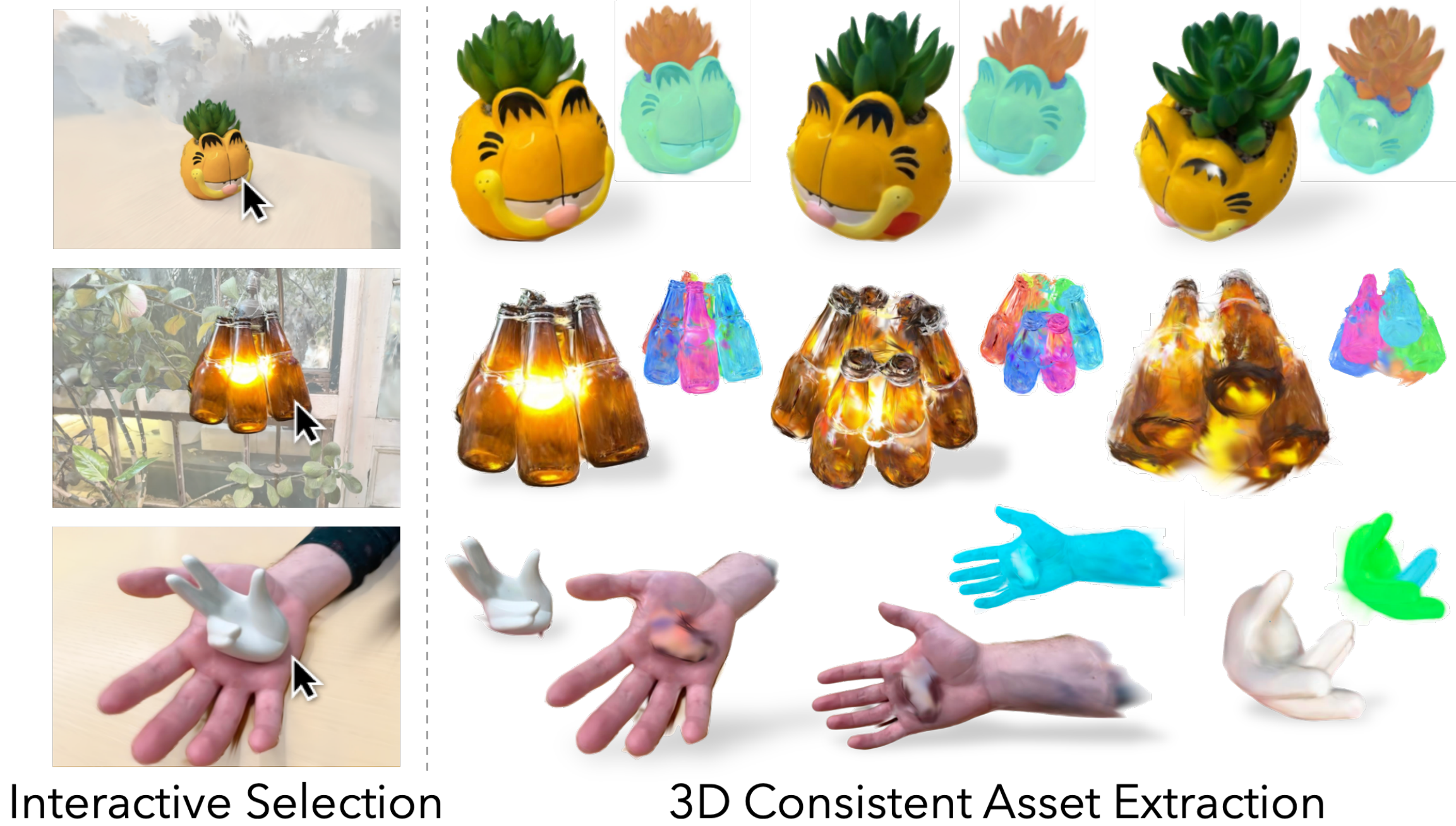}
    \caption{
    \textbf{3D Asset Extraction with Interactive Selection}: Users can interactively select view-consistent 3D groups with \algabbr{} using a click point and a scale.
    }
    \vspace{-1em}
    \label{fig:interactive}
\end{figure}

\begin{figure*}
    \includegraphics[width=\textwidth]{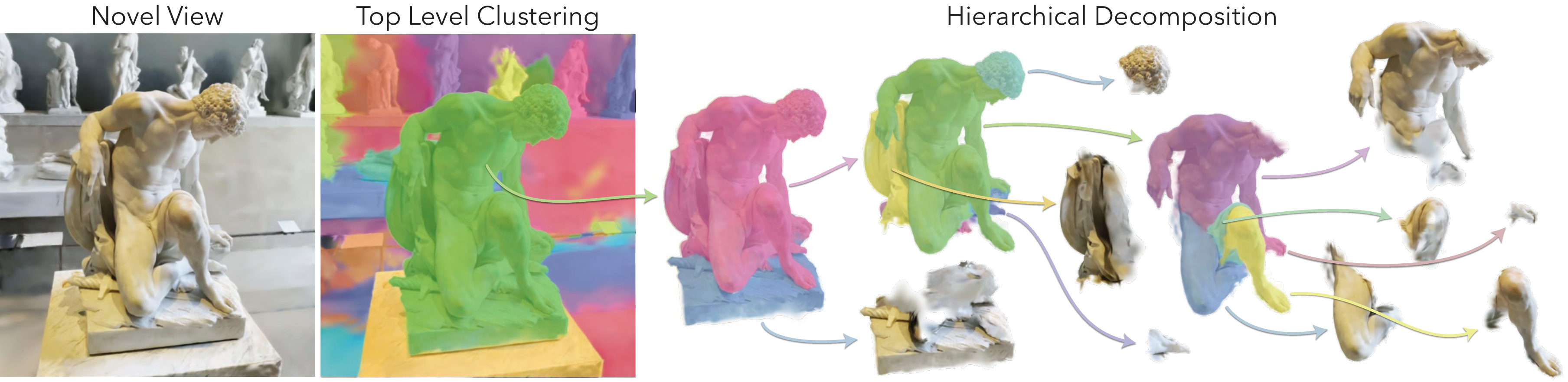}
    \caption{\textbf{3D Decomposition:} \algabbr{} can be recursively queried at decreasing scale to cluster a scene into objects and their subparts.}
    \label{fig:louvre_tree}
    \vspace{-1em}
\end{figure*}
\begin{figure*}
    \includegraphics[width=\textwidth]{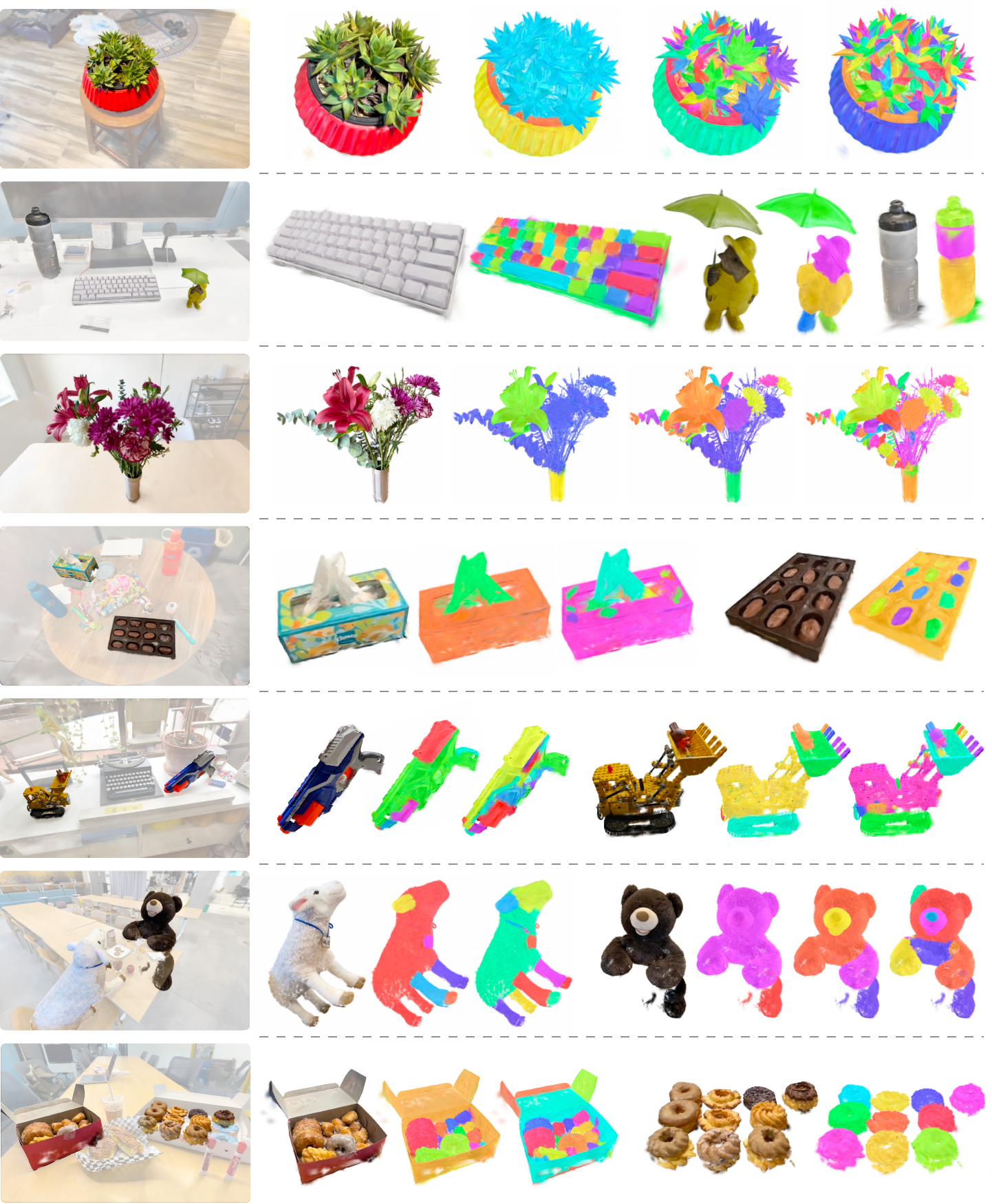}
    \caption{\textbf{Results}: From a \algabbr{} we extract objects from the global scene by selecting top-level clusters, then visualize their local clusters at decreasing scales. \algabbr{} can produce complete 3D object masks, and break these objects into meaningful subparts based on the input masks. We use Gaussian Splats~\cite{kerbl3Dgaussians} to produce these visualizations in 3D. See the Supplemental video for more results.}
    \label{fig:results}
\end{figure*}

\subsubsection{Contrastive Supervision}
The field is supervised with a margin-based contrastive objective, following the definition provided by DrLIM~\cite{hadsell2006dimensionality}. There are two core components of the loss: at a given scale, one which pulls features within the same group to be close, and another which pushes features in different groups apart.

Specifically, consider two rays $r_A, r_B$ sampled from masks $\mathcal{M}_A, \mathcal{M}_B$ within the same training image, with corresponding scales $s_A$ and $s_B$. We can volumetrically render the \grouping{} features along each ray to obtain ray-level features $F_A$ and $F_B$.
If $\mathcal{M}_A = \mathcal{M}_B$, the features are pulled together with L2 distance:
$\mathcal{L}_\text{pull} = ||F_A - F_B||$. 
If $\mathcal{M}_A \neq \mathcal{M}_B$, the features are pushed apart: $\mathcal{L}_\text{push} = \text{ReLU}(m-||F_A - F_B||)$
where $m$ is the lower bound distance, or margin.
Importantly, this loss is only applied among rays sampled from the same image, since masks across different viewpoints have no correspondence.

\subsubsection{Densifying Scale Supervision} \label{sec: hierarchy_losses}

The supervision provided by the previous contrastive losses alone are not sufficient to preserve hierarchy. For example in Fig.~\ref{fig:densification}, although the egg is correctly grouped with the soup at scale 0.22, at a larger scale it fragments apart. 
We hypothesize this grouping instability is because 1) scale supervision is defined sparsely only when a mask exists and 2) nothing imposes containment such that small scale groups remain at larger scales. 
We address these shortcomings here by introducing the following modifications: 

\textbf{Continuous scale supervision} By using 3D mask scales, groups are only defined at discrete values where masks are chosen. This results in large unsupervised regions of scale, as shown at the top of Fig.~\ref{fig:dense_ablation}. We densify scale supervision by augmenting the scale $s$ uniformly randomly between the current mask's scale and the next smallest mask's scale. When a ray's mask is the smallest mask for the given viewpoint, we interpolate between 0 and $s_0$. This ensures continuous scale supervision throughout the field leaving no unsupervised regions.

\textbf{Containment Auxiliary Loss}:
If two rays \( r_1 \) and \( r_2 \) are in the same mask with scale \( s \), then they should also be pulled together at any scale larger than \( s \). Intuitively, two grapes within the same cluster (Fig.~\ref{fig:densification}) are also grouped together at larger scales (e.g., the entire bunch). At each training step, for the rays grouped together at scale $s$, we additionally sample a larger scale $s'>s$ at which the rays are also pulled together. This ensures that affinities at smaller scales are not lost at larger scales.

\subsubsection{Ray and Mask Sampling}
Just like standard NeRF training, we sample rays over which to compute losses. Because \algabbr{} uses a contrastive loss within each train image, naively sampling pixels uniformly during training is inadequate to provide a training signal in each minibatch of rays. To ensure sufficient pairs in each train batch, we first sample N images, and sample M rays within each image. To balance the number of images as well as the number of point pairs for supervision, we sample 16 images and 256 points per image, resulting in 4096 samples per train iteration. 

For each ray sampled, we must also choose a mask to use as the group label for the train step in question. To do this, we retain a mapping from pixels to mask labels throughout training, and at each train step randomly select a mask for each ray from its corresponding list of masks. There are two important caveats in this sampling process: 
\noindent\textbf{1)} The probability a mask is chosen is weighted inversely with the log of the mask's 2D pixel area. This prevents large scales from dominating the sampling process, since larger masks can be chosen via more pixels. 
\noindent\textbf{2)} During mask selection we coordinate the random scale chosen across rays in the same image to increase the probability of positive pairs. To do this, we sample a single value between 0 and 1 per image, and index into each pixel's mask probability CDF with the same value, ensuring pixels which land within the same group are assigned the same mask. Otherwise, the loss is dominated by pushing forces which destabilize training.

\subsection{Implementation Details}
The method is built in Nerfstudio~\cite{tancik2023nerfstudio} on top of the Nerfacto model by defining a separate output head for the grouping field. The grouping field is represented with a hashgrid~\cite{muller2022instant} with 24 layers and a feature dimension of 2 per layer, and a 4-layer MLP with 256 neurons and ReLU activation which takes in scale as an extra input concatenated with hashgrid feature. We cap scale at $2\times$ the extent of cameras, and normalize the scale input to the MLP using sklearn's quantile transform on the distribution of computed 3D mask scales (Sec~\ref{method:maskgen}). Output embeddings are $d=256$ dimensions. Gradients from the affinity features do not affect the RGB outputs from NeRF, as these representations share no weights or gradients.

We begin training the grouping field after 2000 steps of NeRF optimization, giving geometry time to converge. In addition, to speed training we first volumetrically render the hash value, then use it as input to the MLP to obtain a ray feature. With this deferred rendering, the same ray can be queried at different scales with only one extra MLP call. We normalize the result of volume rendering to unit norm before inputting to the MLP, and for point-wise queries, the individual hashgrid value is normalized. Preprocessing SAM masks takes around 3-10 minutes, followed by about 20 minutes for training on a GTX 4090.

\section{Hierarchical Decomposition} \label{sec:hierarchical}
Once we have optimized a scale-conditioned affinity, \algabbr{} generates a hierarchy of 3D groups, organized in a tree such that each node is broken into potential sub-groups. To do this we recursively cluster groups by decreasing the scale for affinity, using {HDBSCAN}~\cite{mcinnes2017hdbscan}, a density based clustering algorithm which does not require a prior on number of clusters. This clustering process can be done in 2D on volumetrically rendered features in an image which yields masks, or in 3D across points to yield pointclouds. See Fig.~\ref{fig:louvre_tree} for a visualization of scene decomposition.

\noindent\textbf{Initialization}: First, to initialize the hierarchy, we first globally cluster features at a large scale $s_\text{max}$, which we set to $1.0$ for all experiments, corresponding to the extent of the input cameras' positions. These clusters form the top-level nodes in the scene decomposition.

\noindent\textbf{Recursive Clustering}: Next, to produce a hierarchical tree of scene nodes, we iteratively reduce scale by a fixed epsilon (we use 0.05), running HDBSCAN on each leaf node. If HDBSCAN returns more than one cluster for a given node, we add those clusters as children and recurse. This continues until we reach scale 0, at which point the procedure terminates, returning the current tree.

\begin{figure}
    \includegraphics[width=\linewidth]{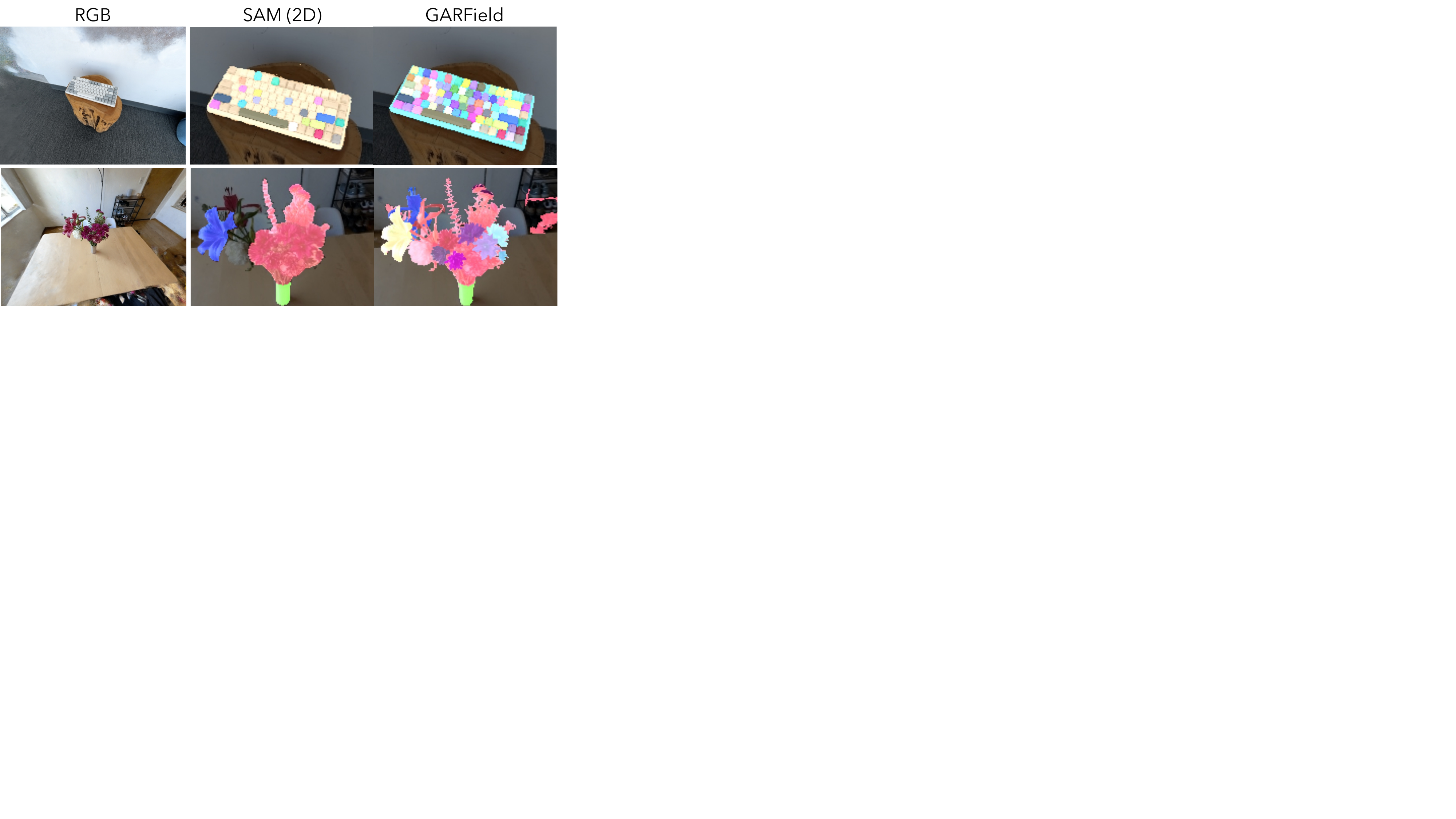}
    \caption{\textbf{Segment-Anything~\cite{kirillov2023segment} vs. \algabbr{}}: SAM's automatic mask generator may encounter difficulty recalling all masks from a given viewpoint, especially when there are clusters of small masks and the camera is far away from the object. In contrast, \algabbr{}'s \grouping{} field incorporates masks from multiple viewpoints in 3D. \vspace{-1em}}
    \label{fig:sam_comparison}
\end{figure}

\section{Experiments}
We assess \algabbr{}'s ability to decompose in-the-wild 3D scenes into hierarchical groups which vary widely in size and semantics. 
Existing 3D scan datasets tend to focus on object-level scans~\cite{downs2022google,mo2019partnet}, are simulated~\cite{bhalgat2023contrastive}, or contain primarily indoor household scenes~\cite{scannet}. To evaluate \algabbr{}, we instead use a wide variety of indoor and outdoor scenes from the Nerfstudio and LERF datasets, as well as additional captures for this paper. We experiment on scenes which possess significant object hierarchy, testing the decomposition ability of \algabbr{}. We provide qualitative results in Fig.~\ref{fig:method} and Fig.~\ref{fig:louvre_tree}, and quantitatively evaluate by annotating ground truth masks on select scenes, a full list of which are in the Supplement.

\begin{figure}
    \includegraphics[width=\linewidth]{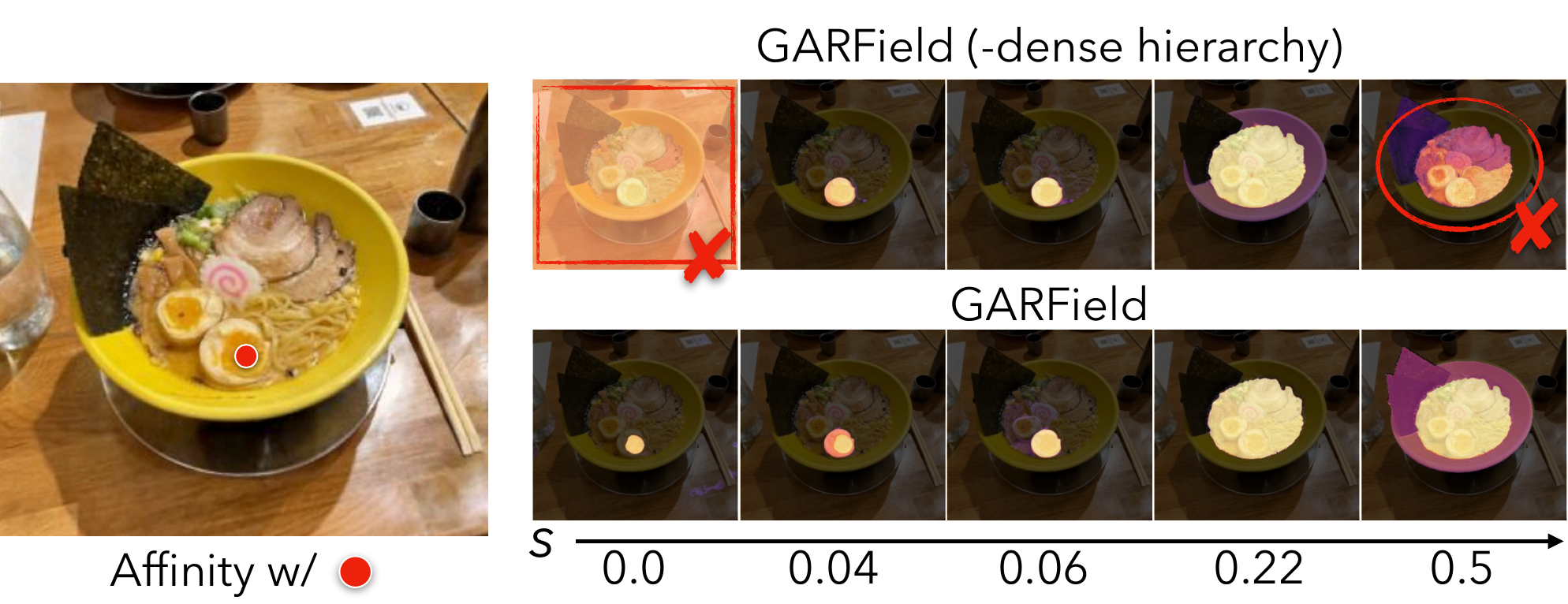}
    \caption{\textbf{Ablation}: Without dense hierarchy supervision, points may have inconsistent affinities across scales. There may be 1) spurious large affinities at unsupervised scales or 2) unexpected drops in affinity at larger scales.}
    \vspace{-1em}
    \label{fig:dense_ablation}
\end{figure} 
\subsection{Qualitative Scene Decomposition}
We use Gaussian Splatting~\cite{kerbl3Dgaussians} to visualize the decomposition by querying \algabbr{}'s affinity field at gaussian centers. We do this because gaussian splats are easier to segment in 3D compared to NeRFs. See the Supplement for a full description of the pipeline. All renderings are of complete 3D models, not segmentations of 2D image views.

We visualize two types of hierarchical clustering results. In Fig.~\ref{fig:results} we globally cluster the scene at a hand selected coarse scale, then from these scene-wide clusters we select groups corresponding to few objects and further decompose them into subgroups. We visualize clusters obtained at successively decreasing scales, which increases the granularity of groups. \algabbr{} achieves high-fidelity 3D groupings across a wide range of scenes and objects, from man-made objects -- such as keyboards, where each key is considered a group at a small scale, to the parts of the NERF gun and the Lego bulldozer -- to complex natural objects like plants, where it can group individual flowers as well as their petal and leaves.
By varying the scale, one can separate objects at different levels, for instance the succulent from its pot versus each individual leaf (first row), or identifying the bunny toy in the bulldozer's scooper, which is further grouped into its shirt, ears, and head (fifth row, right). See Fig.~\ref{fig:scene-wide} for select scene-wide cluster visualizations.

In Fig.~\ref{fig:louvre_tree} we visualize a tree decomposition produced by the method described in Sec.~\ref{sec:hierarchical}. We first show the global clustering at a top level node, from which we select the central statue to illustrate the tree decomposition. Arrows denote children in the hierarchy, illustrating how the statues decomposes gradually all the way down to its hair, legs, torso, etc. See the Supplement for more tree visualizations.

\begin{table}
\centering
\hspace{2em}
\begin{tabular}{l@{\extracolsep{4pt}}r@{\extracolsep{4pt}}rr@{\extracolsep{4pt}}rr@{\extracolsep{4pt}}r}
\toprule
& \multicolumn{2}{c}{Fine} & \multicolumn{2}{c}{Medium} & \multicolumn{2}{c}{Coarse} \\
Scene & SAM & Ours & SAM & Ours & SAM & Ours \\
\midrule
teatime & 81.6 & \textbf{92.7} & 97.3 & \textbf{97.9} & - & - \\
bouquet & 17.4 & \textbf{76.0} & 73.5 & \textbf{81.6} & 76.1 & \textbf{85.4} \\
keyboard & 65.3 & \textbf{88.8} & 73.6 & \textbf{98.4} & - & - \\
ramen & 53.3 & \textbf{79.2} & 74.7 & \textbf{90.7} & 92.6 & \textbf{95.5} \\
living\_room & 85.3 & \textbf{90.5} & 74.2 & \textbf{80.7} & 88.6 & \textbf{94.4} \\
\bottomrule %
\end{tabular}
\vspace{-0.5em}
\caption{\textbf{3D Completeness.} We report mIOU of scene annotations for a single point with up to three levels of hierarchy. SAM struggles to produce view-consistent fine groups compared to \algabbr{}.}
\label{tab:scene_comparison}
\vspace{0.5em}
\begin{tabular}{lcccc}
\toprule
\multirow{2}{*}{Scene} & \multirow{2}{*}{SAM~\cite{kirillov2023segment}} & Ours & Ours & \multirow{2}{*}{Ours} \\ 
 &  & (-scale) & (-dense) & \\
\midrule
ramen & 74.9 & 64.1 & 74.1 & \textbf{85.6 }\\
teatime & 64.9 & 67.7 & 66.1 & \textbf{86.6} \\
keyboard & 23.2 & 57.6 & 73.1 & \textbf{77.9} \\
bouquet & 34.4 & 49.8 & 72.9 & \textbf{76.4} \\
living\_room & 59.6 & 49.7 & 62.1 & \textbf{76.6} \\ 
\bottomrule %
\end{tabular}
\caption{\textbf{Hierarchical Grouping Recall:} We report mIOU against human annotations of multi-scale groups of different objects.}
\label{tab:recall}
\vspace{-1em}
\end{table}

\subsection{Quantitative Hierarchy}
We quantitatively evaluate our approach against annotated images using two metrics: the first measuring view consistency against annotations from multiple views and the second measuring recall of various hierarchical masks via mIOU against ground truth human annotations.

\noindent\textbf{3D Completeness:} For downstream tasks it is useful for groups to correspond to complete 3D objects, for example groups that contain an entire object rather than just one of its sides.
Though \algabbr{} always produces view-consistent groups by construction, it may not necessarily contain complete objects. 
We evaluate for completeness by checking that an entire 3D object is grouped together across a range of viewpoints. To do this, on 5 scenes we choose a 3D point to be projected into 3 different viewpoints, and label 3 corresponding view-consistent ground truth masks containing that point at coarse, medium, and fine levels. At these points we mine multiple masks from \algabbr{} across multiple scales at 0.05 increments, where at each scale a mask is obtained based on feature similarity thresholded at 0.9.
We also compare against SAM by clicking the point in the image and taking all 3 masks. We report the maximum mIOU computed over all candidate masks for both methods. 

Results are shown in Table~\ref{tab:scene_comparison}. \algabbr{} produces more complete 3D masks than SAM across viewpoints, resulting in higher mIOU with multi-view human annotations of objects. This effect is especially apparent at the most granular level, where from certain perspectives SAM struggles to produce fine groups, like the keyboard keys from afar in Fig.~\ref{fig:sam_comparison}. See the Supplement for figures of comparisons and visualization of the groundtruth masks.

\noindent\textbf{Hierarchical Grouping Recall:} Here we measure \algabbr{}'s ability to recall groups at multiple granularities. Across 5 scenes, we choose one \textit{novel} viewpoint and label up to 3 ground truth hierarchical groups for 1-2 objects. \algabbr{} outputs a set of masks as described in Section~\ref{sec:hierarchical} by clustering image-space features, outputting one mask per tree node. We compare against SAM's automatic mask generation by keeping all output masks.
We ablate \algabbr{} in two ways: \algabbr{} (-scale) removes scale-conditioning; and \algabbr{} (-hierarchy) removes the densified supervision in Sec.~\ref{sec: hierarchy_losses}.

In Table ~\ref{tab:recall} we report mIOU of the ground truth mask with the highest overlap, either from the set of SAM masks or the tree generated by \algabbr{}. Because \algabbr{} has fused groups from multiple perspectives, it results in higher fidelity groupings than any single view of SAM, leading to higher mIOU with annotations. Our ablations show that scale conditioning and scale densification is necessary for 
high quality groupings. Fig.~\ref{fig:dense_ablation} illustrates affinity degrading at higher scale with naive supervision.

\section{Limitations}
\algabbr{} at its core is distilling outputs from a 2D mask generator, so if the masks fail to contain a desired group, this will not emerge in 3D. Regions with uneven viewpoints can suffer from artificial group boundaries, for example if an object is only viewed from close up, it may never be grouped together because no input view contained it in full. We handle group ambiguity using physical size, but there could be multiple groupings within a single scale.
For example, conflicts may happen with objects contained in a container because the container with and without the object can have the same scale. Future work could consider other ways to resolve grouping ambiguity such as affordances. 
Another consequence of scale-conditioning is that object parts of different sizes branch off the tree separately rather than all at once: multiple objects on the same table may appear at different levels of the tree. The tree generation in this work is a naive greedy algorithm, which can result in spurious small groups at deeper levels, as seen in the trees in the Supplement. Future work may explore more sophisticated ways of hierarchical clustering.
\begin{figure}
    \centering
    \includegraphics[width=\linewidth]{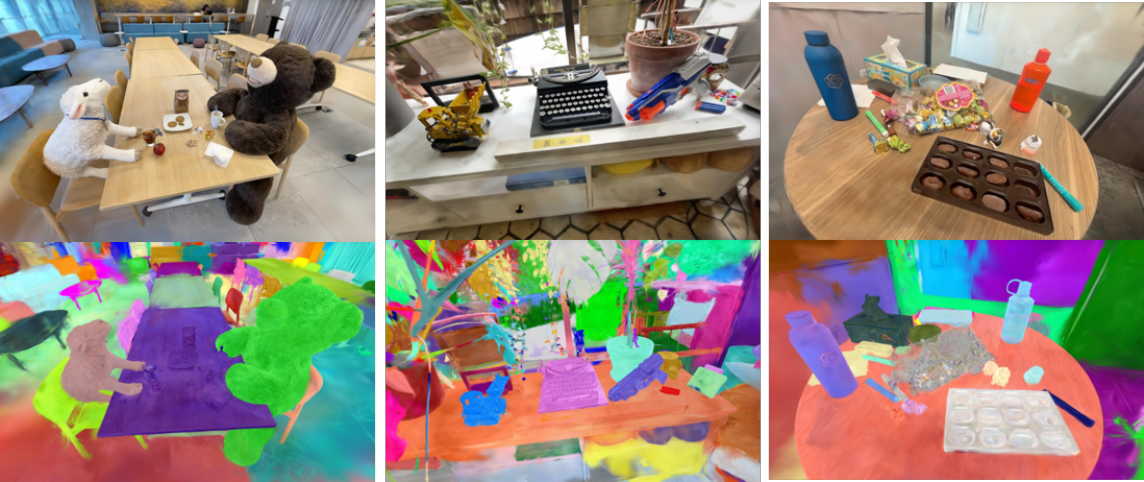}
    \caption{\textbf{Scene-Wide Clustering} visualizations for selected scenes from Fig.~\ref{fig:results}.}
    \label{fig:scene-wide}
    \vspace{-1em}
\end{figure}

\section{Conclusion}
We present \algabbr{}, a method for distilling multi-level masks into a dense scale-conditioned affinity field for hierarchical 3D scene decomposition. By leveraging scale-conditioning, the affinity field can learn meaningful groups from conflicting 2D group inputs and break apart the scene at multiple different levels, which can be used for extracting assets at a multitude of granularities. \algabbr{} could have applications for tasks that require multi-level groupings like robotics, dynamic scene reconstruction, or scene editing.

\section{Acknowledgements}
This project was funded in part by NSF:CNS-2235013 and DARPA Contract No. HR001123C0021. Chung Min and Justin are supported in part by the NSF Graduate Research Fellowship Program under Grant No. DGE 2146752. Any opinions, findings, and conclusions or recommendations expressed in this material are those of the author(s) and do not necessarily reflect the views of the National Science Foundation.

\newpage
{
    \small
    \bibliographystyle{ieeenat_fullname}
    \bibliography{main}
}

\clearpage
\appendix
\setcounter{page}{1}
\maketitlesupplementary

\section{Additional Results}
We show additional figures and videos using \algabbr{} for 1) hierarchical decomposition, 2) global clustering, and 3) interactive selection. All video visualizations use Gaussian Splatting~\cite{kerbl3Dgaussians}, as described below. 

\subsection{Gaussian Splat Visualizations}
We use Gaussian Splatting~\cite{kerbl3Dgaussians} to emphasize the 3D nature of \algabbr{} and its applications for 3D group extraction. Here, for simplicity, we do not optimize \algabbr{} directly with gaussians. Instead, we train a NeRF-based \algabbr{} and a Gaussian Splatting model separately. Then, we assign an affinity feature to every gaussian by querying the feature field at the gaussian's center point. We use these features to manipulate the 3D scene, \eg clustering, selection, and filtering. All implementation described here will be made public. To visualize clusters in 3D, we override each gaussian's color parameters to the RGB color of the colormap.

\subsection{3D Hierarchical Decomposition}
In the main text, we visualized hand-picked nodes from the resulting hierarchy in Main Paper Fig.~\ref{fig:louvre_tree}. Here, we exhaustively visualize entire subtrees of selected scenes by selecting the primary region of interest (i.e. desk, dozer, bouquet). 
\subsubsection{Full Tree Visualizations}
\begin{figure*}
    \centering
    \includegraphics[height=.97\textheight]{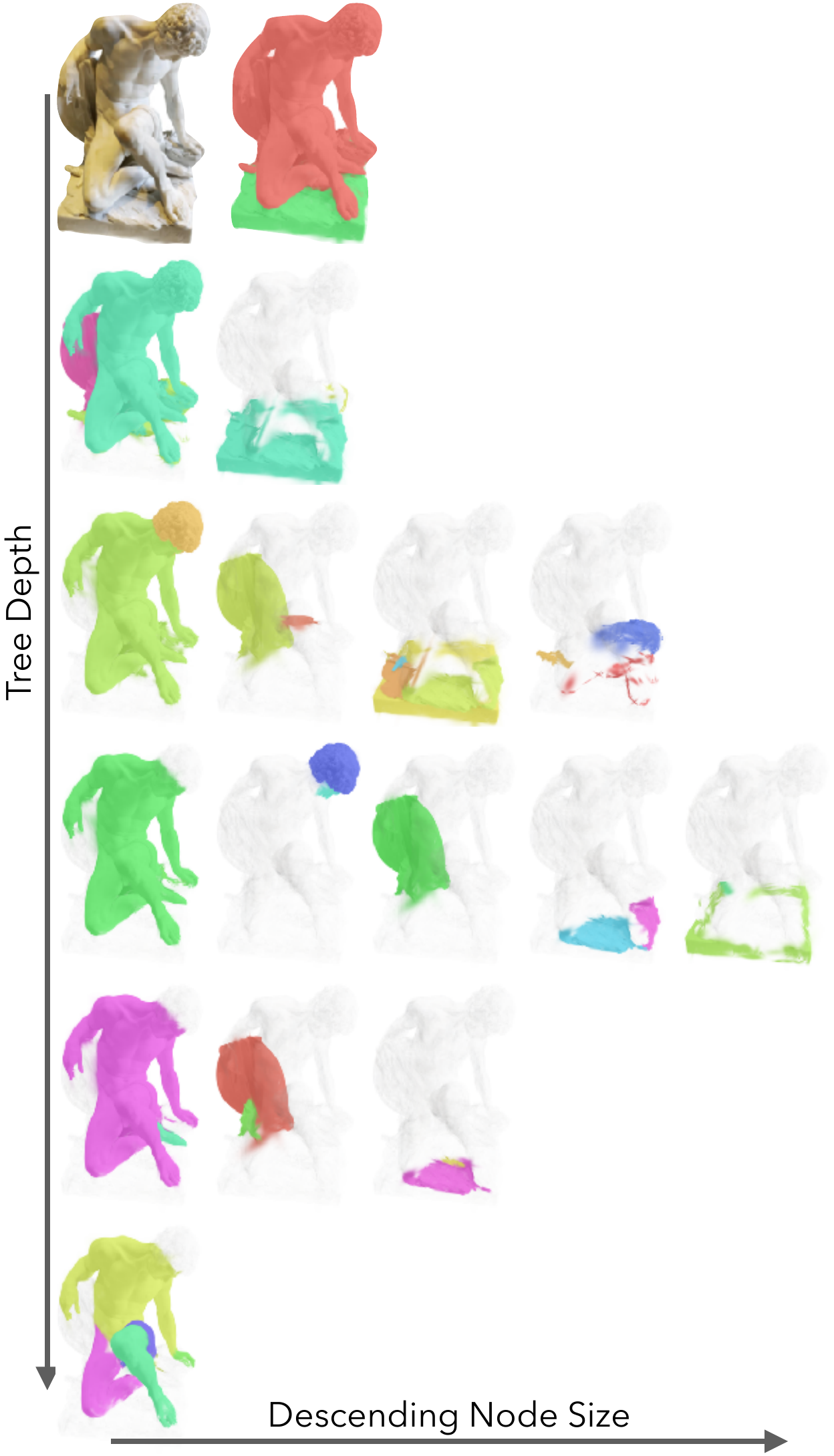}
    \caption{\textbf{Complete Tree}: A complete visualization of all layers and all nodes in the tree from Fig.~\ref{fig:louvre_tree}. Colors illustrate different clusters within each node, and each row visualizes all the nodes at a given depth in the tree, sorted by size.}
    \label{fig:louvre_all_nodes}
\end{figure*}

In Fig.~\ref{fig:louvre_all_nodes} and in provided videos we visualize each layer of the resulting tree organized by node depth in different rows. Each node is shown colorized by the number of internal clusters, with the remainder of the tree drawn with low opacity to give context. Note that nodes at the same level do not necessarily correspond to the same scale because intermediate nodes are pruned.

One can see how each part is recursively broken into subparts in lower layers of the tree, for example the statue gets broken into the base and rest of the statue, followed by shield, torso, hair, and etc.
Note how some nodes can contain noise or partial clusters, for example the third row, last node of Fig.~\ref{fig:louvre_all_nodes}, where the red cluster is a spurious cluster which more suitably belongs to the base of the statue in the prior tree level. Artifacts such as this can happen as a result of our greedy tree building approach, and might be addressed with a more sophisticated tree construction algorithm. Videos of trees showcase the view-consistency of 3D scene decomposition, with whole objects being clustered together like the bear or dozer, which can then be broken into coherent subparts. The lowest levels of the tree contain very fine details such as petals of flowers, or hooves of the sheep. 

These exhaustive tree visualizations also exhibit limitations, such as spurious background points being grouped together with the object of interest, a behavior which could be remedied by more strongly taking geometric proximity into account when constructing the tree. Another failure mode is that when view coverage is insufficient, different sides of the same object can be grouped separately. For example, in rows 3 and 4 of Fig.~\ref{fig:louvre_all_nodes} the two sides of the statue's face are grouped differently. 

\subsubsection{Compressed Tree Visualizations}
We additionally provide videos of compressed trees, where each layer of the tree is merged into one visual by distinctly coloring all clusters. Leaf nodes at one layer are further propagated to deeper layers of the tree to visualize all clusters at the lowest level, corresponding to the most granular decomposition. Though these visualizations do not show hierarchy because they merge all nodes, they illustrate how lower layers of the scene decomposition correspond to semantically meaningful high granularity and higher levels correspond to coarser granularities.
\subsection{Multi-Scale Clustering}
We provide video versions of Main Paper Fig.~\ref{fig:results} to showcase the view-consistency of the results shown in the images. These videos first show the global clustering of the scene, followed by video renderings of sub-object clusters. 
\subsection{Global Clustering}
\begin{figure*}
    \centering
    \includegraphics[width=\linewidth]{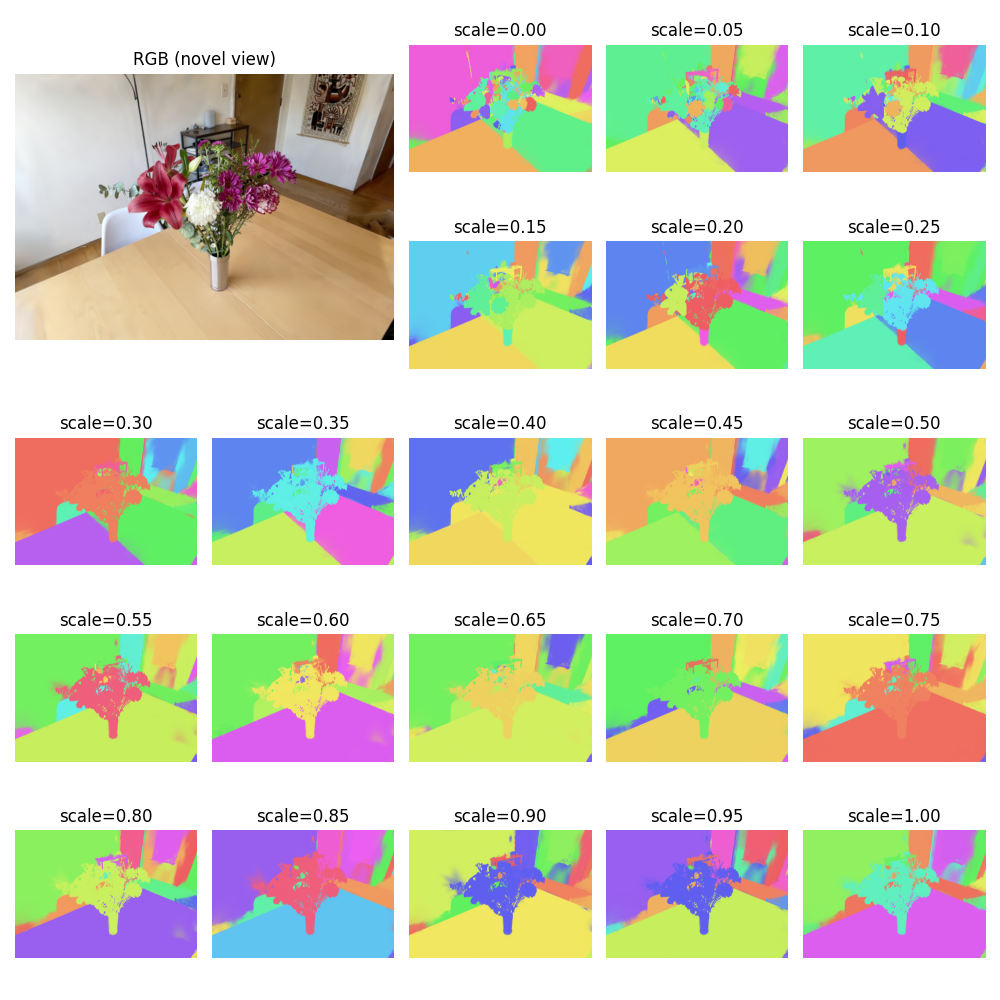}
    \caption{\textbf{Global Clustering Results (``Bouquet")}:  Global clusters at smaller scales $(s=0)$ distinguish between different sections of the bouquet, as well as the two halves of the table. At a larger scale, the bouquet and table are considered whole.}
    \label{fig:global_bouquet}
\end{figure*}

\begin{figure*}
    \centering
    \includegraphics[width=\linewidth]{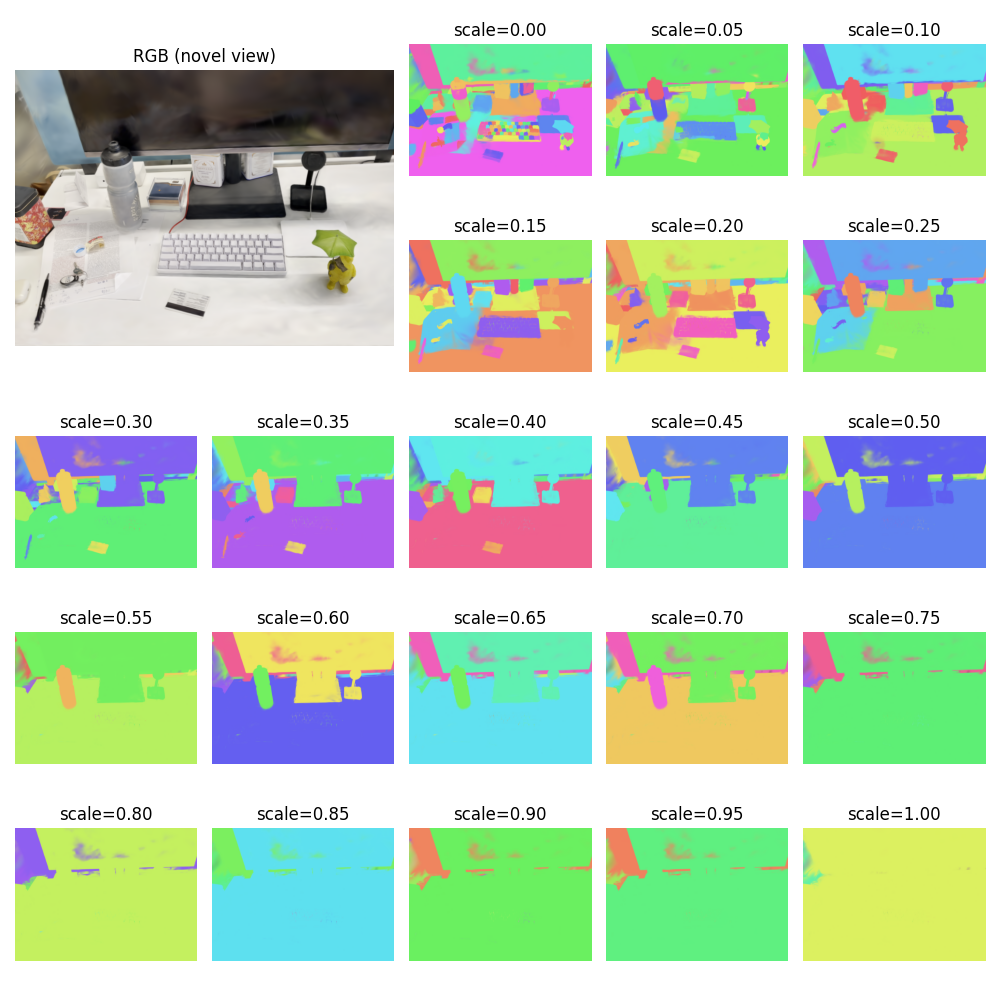}
    \caption{\textbf{Global Clustering Results (``Desk")}: At larger scales $(s=0.5)$, the desk is grouped together with the clutter on it \eg keyboard, card, bird figurine).}
    \label{fig:global_desk}
\end{figure*}

\begin{figure*}
    \centering
    \includegraphics[width=\linewidth]{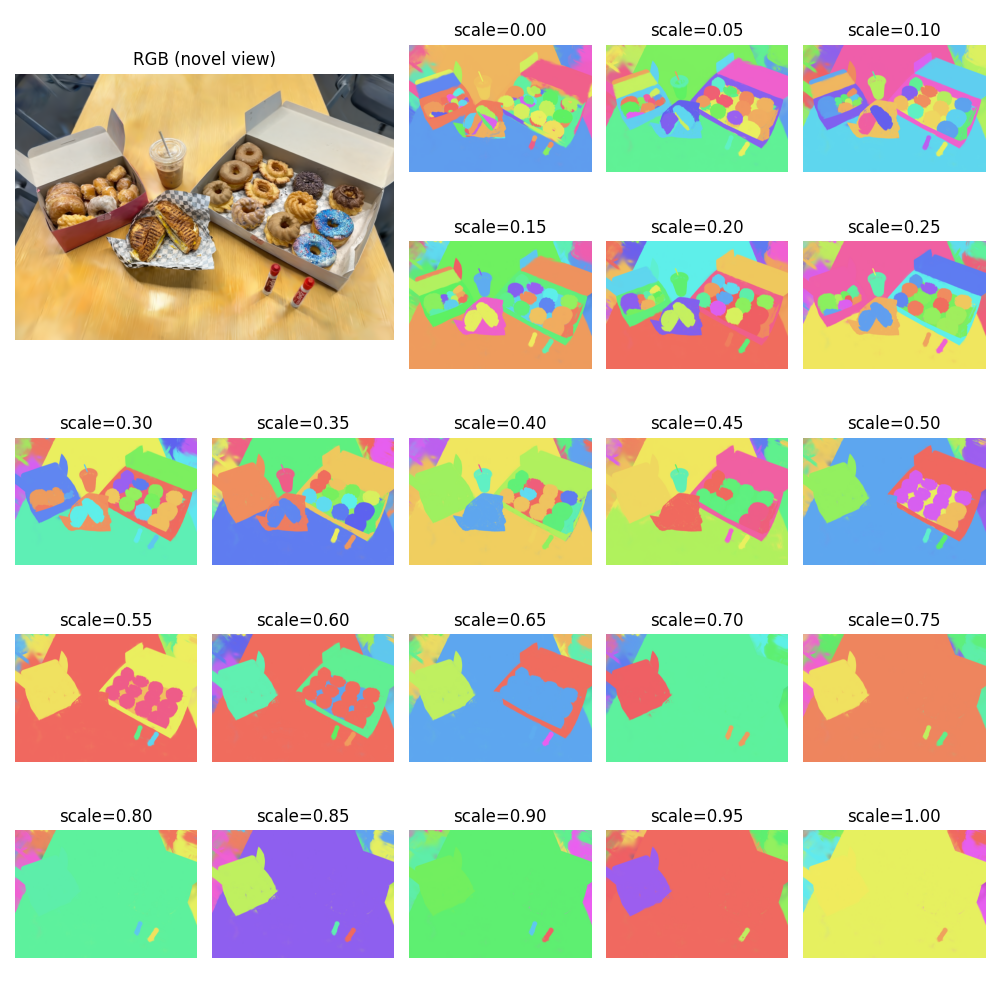}
    \caption{\textbf{Global Clustering Results (``Donuts")}: At a very small scale $(s=0.0)$, \algabbr{} can distinguish between different pieces of the breakfast sandwich in the middle of the scene. As scale increases, its grouping shifts quite noticably --- into its two halves, or the full sandwich with the checkerboard packaging.}
    \label{fig:global_donuts}
\end{figure*}

\begin{figure*}
    \centering
    \includegraphics[width=\linewidth]{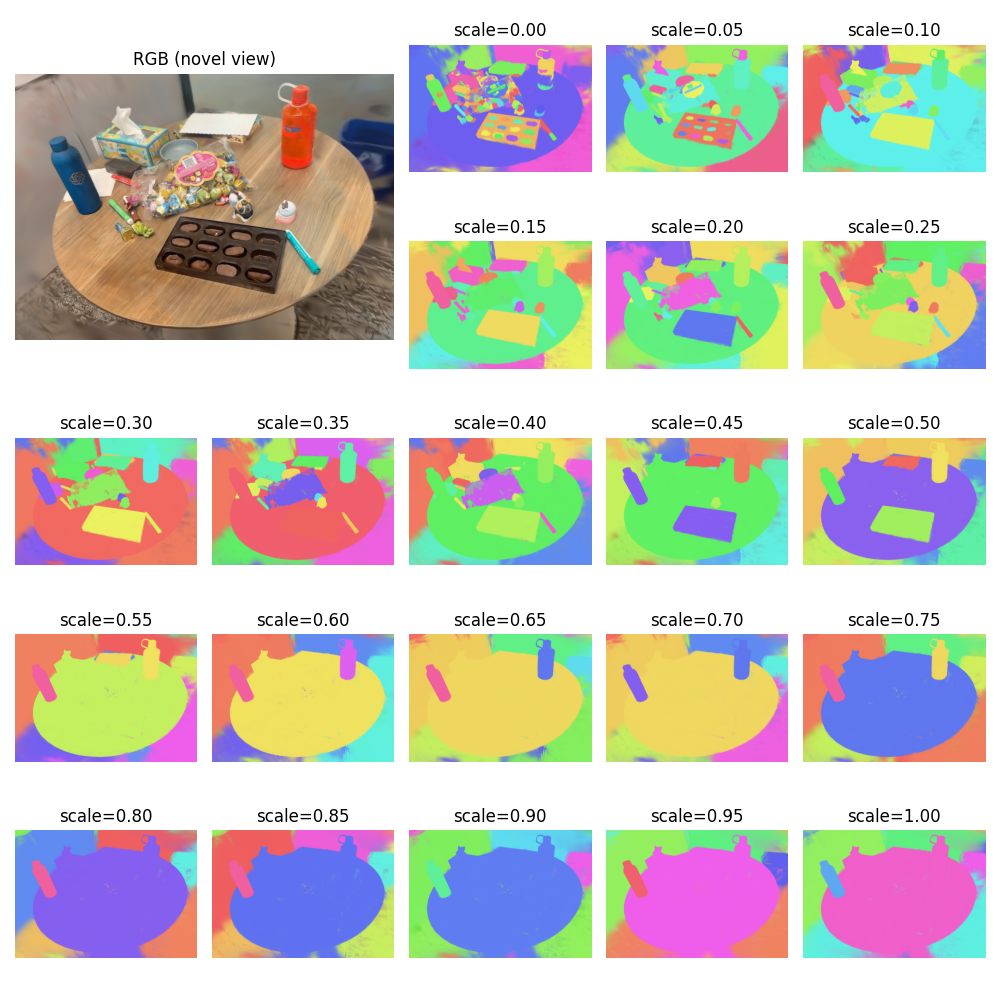}
    \caption{\textbf{Global Clustering Results (``Table")}: At the smallest scale $(s=0.0)$, the global clusters highlight parts of objects \eg labels on water bottles, pieces of chocolate.}
    \label{fig:global_table}
\end{figure*}

\begin{figure*}
    \centering
    \includegraphics[width=\linewidth]{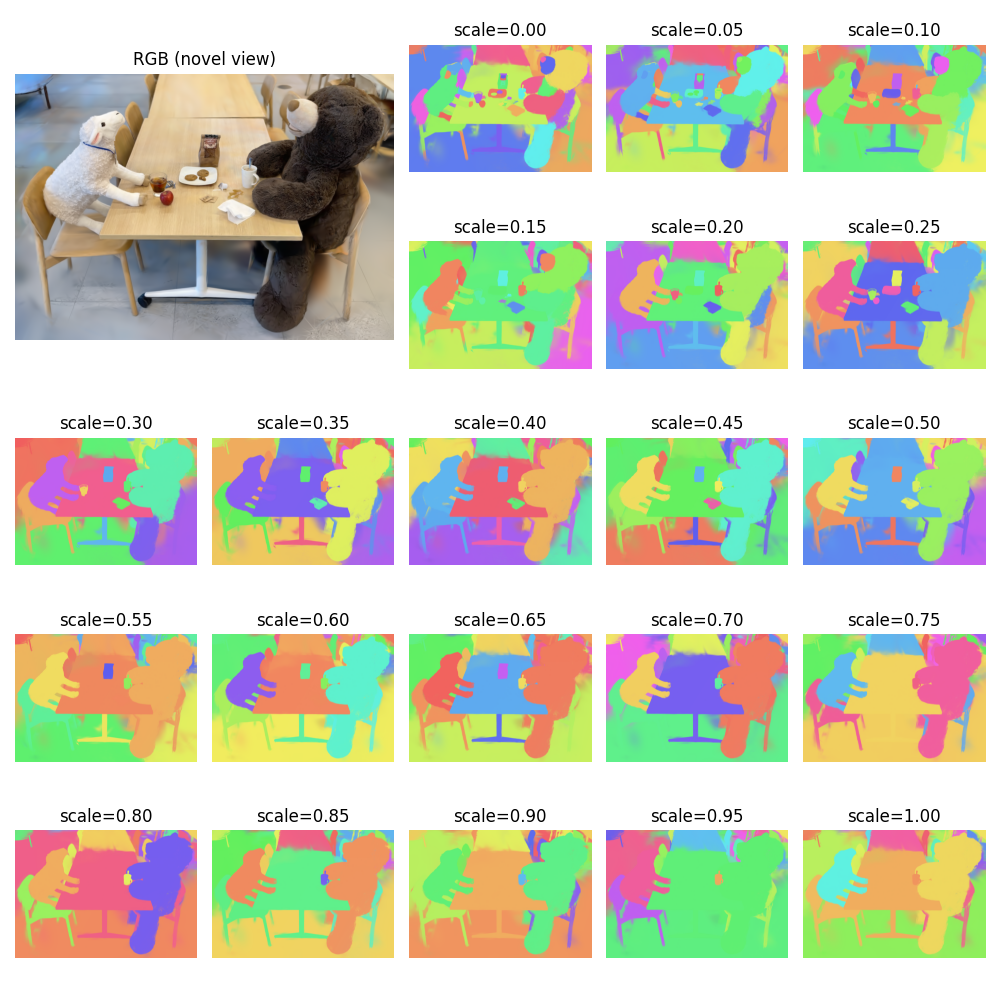}
    \caption{\textbf{Global Clustering Results (``Teatime")}: The food, utensils, and the table are included in different clusters at small scales, and the same cluster at larger scales. Parts of the stuffed animals (\eg sheep hooves, bear nose) can also be seen at $s=0.0$.}
    \label{fig:global_teatime}
\end{figure*}

\begin{figure*}
    \centering
    \includegraphics[width=\linewidth]{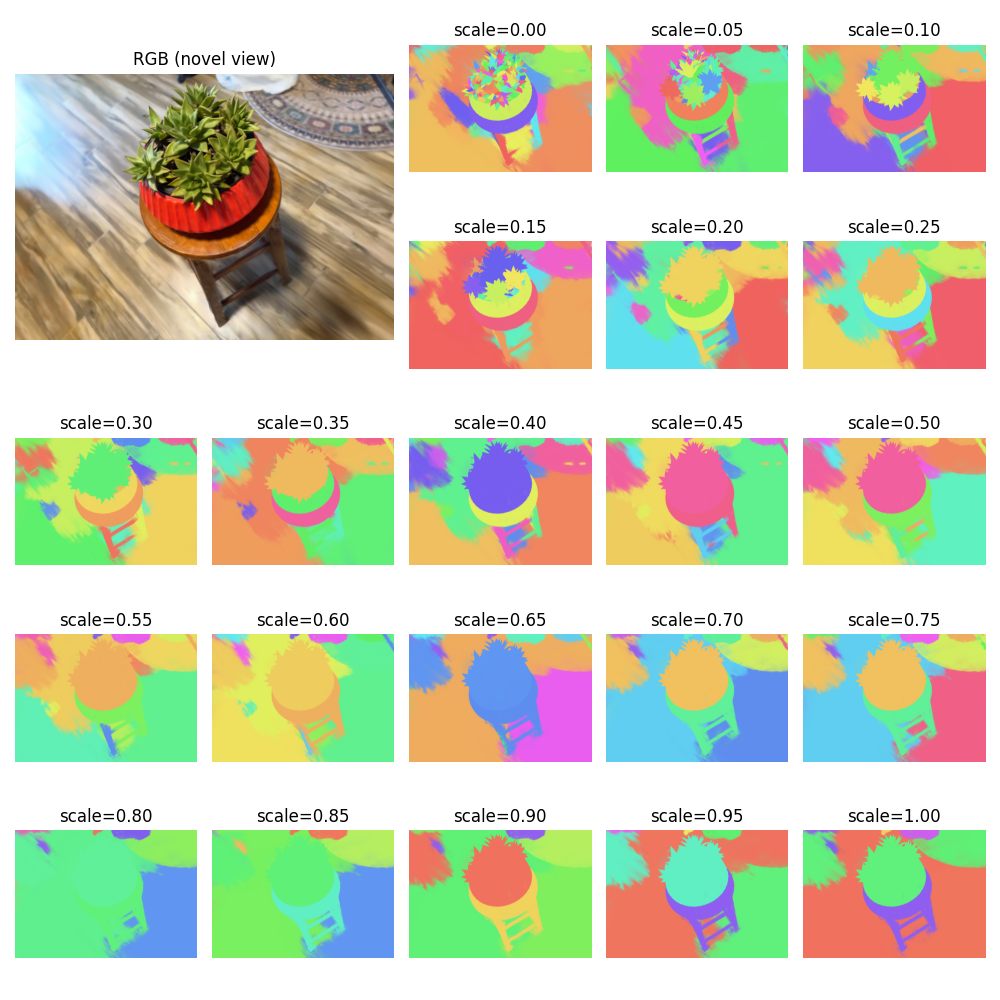}
    \caption{\textbf{Global Clustering Results (``Succulent")}: Global clusters at smaller scales ($s=0.0$) distinguish between fine features like succulent leaves, while they are considered a single group at larger scales ($s=1.0$).}
    \label{fig:global_succulents}
\end{figure*}

\begin{figure*}
    \centering
    \includegraphics[width=\linewidth]{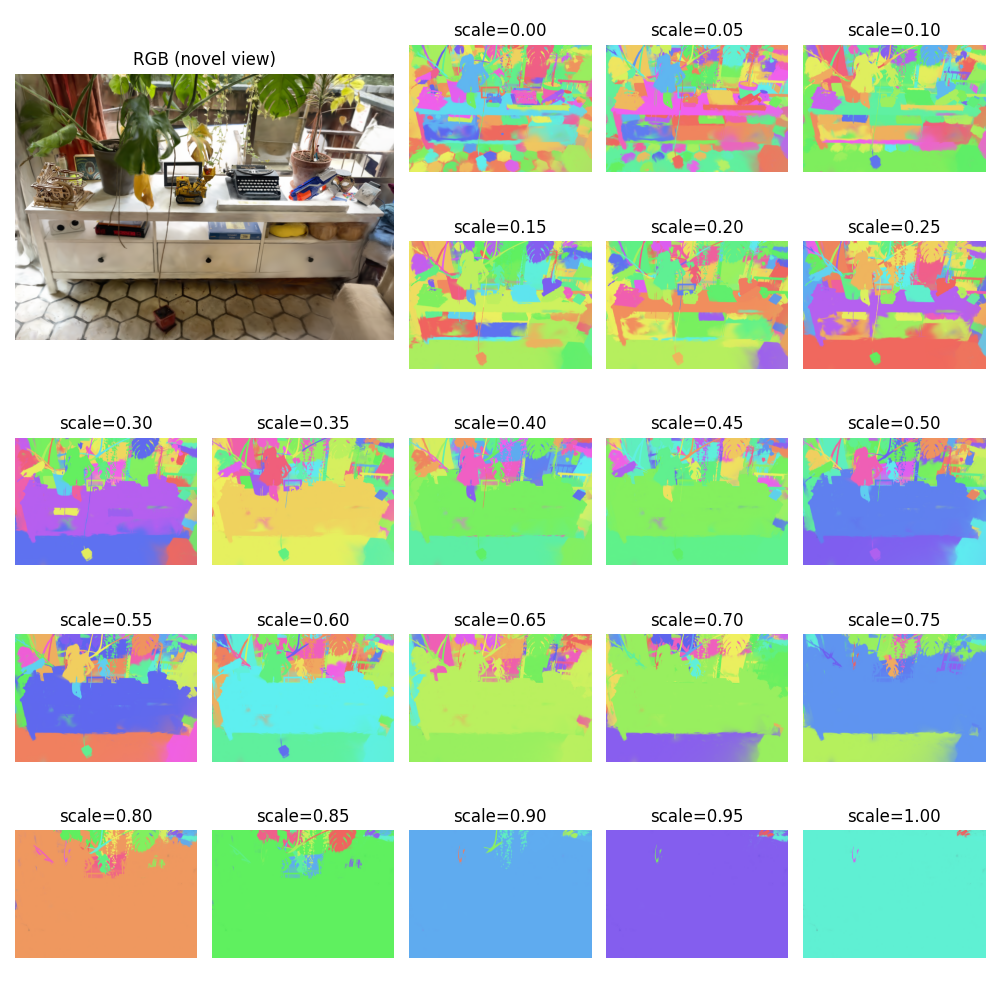}
    \caption{\textbf{Global Clustering Results (``Living Room"}: The individual hexagonal tiles on the floor may be grouped separately ($s=0.0$) or together $(s=0.5)$. }
    \label{fig:global_livingroom}
\end{figure*}

To emphasize that \algabbr{} can model scene-level groupings, we cluster \algabbr{} features globally \ie all gaussians in a scene. Figures~\ref{fig:global_bouquet} through \ref{fig:global_livingroom} show all scenes in Fig.~\ref{fig:results} globally clustered at scales 0 to 1, at increments of 0.05. 

We also include a video where the excavator scene in Main Paper Fig.~\ref{fig:teaser} is globally clustered at three distinct scales. We find that \algabbr{} successfully groups together large group in the backgrounds, like the road or bushes on the sidewalk.

\subsection{Interactive Selection}
People can use clicks to interact with \algabbr{} and extract groups of different sizes, as shown in Fig.~\ref{fig:interactive} of the main paper. User clicks are transformed into 3D points using projective geometry (visualized with a red sphere in the video). At a given scale, we select a set of 3D gaussians based on their affinity with the selected point. To retrieve multiple groups, we query \algabbr{} across a range of scales and merge groups with large overlap. In the video, a user can extract the excavator, crane, and scooper from Fig.~\ref{fig:teaser} with a single click.

\section{Experiment details}
\begin{figure}[h]
    \centering
    \includegraphics[width=\linewidth]{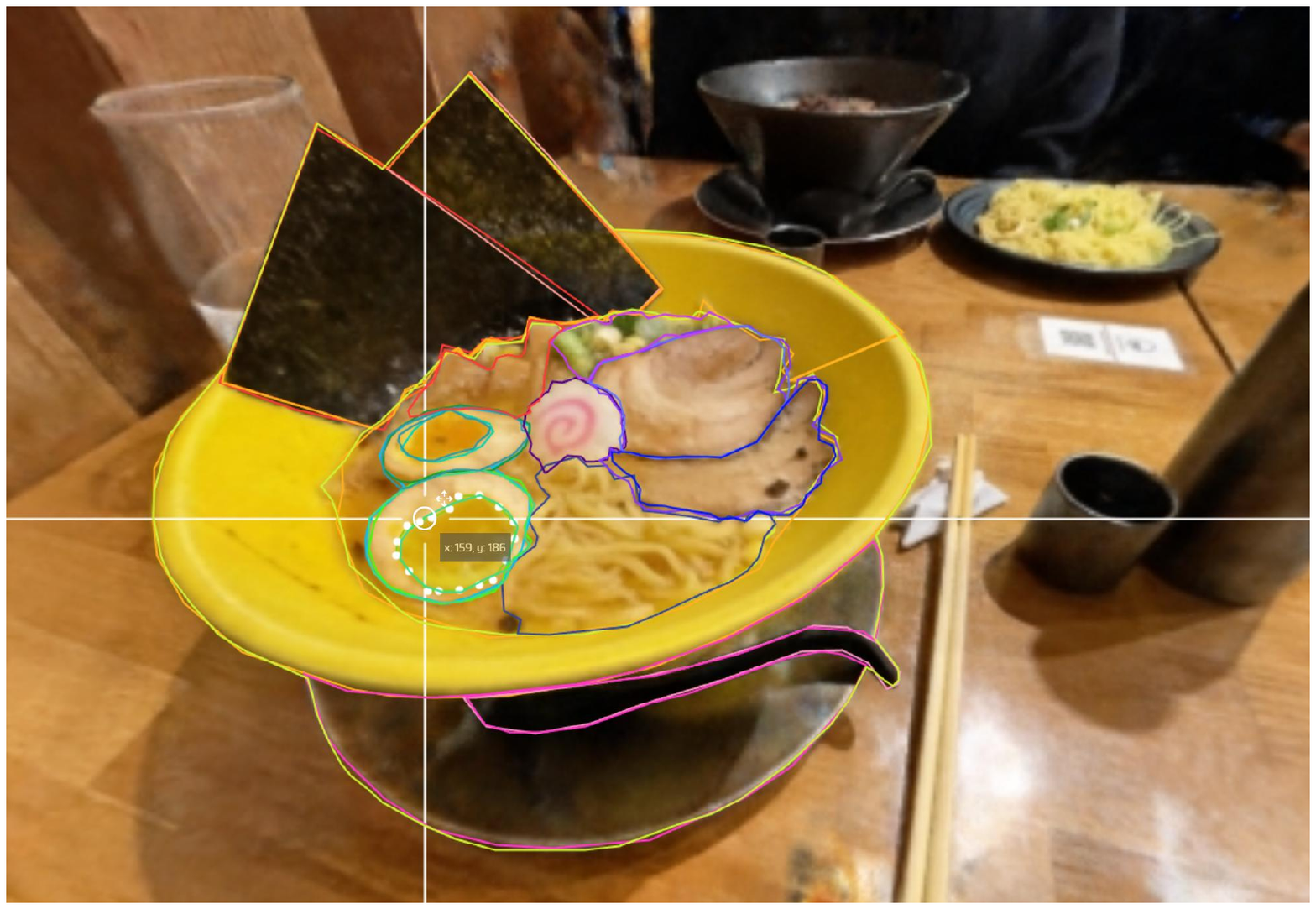}
    \caption{\textbf{Masks for 3D Completeness Experiments}: Overlapping masks (\textit{egg}, \textit{noodles}, \textit{nori} masks inside \textit{ramen} mask) model the desired hierarchical groupings. We labeled these polygonal masks using `Make Sense'~\cite{make-sense}, an online tool for mask annotation.}
    \label{fig: label_ground_truth}
\end{figure}
\subsection{Hierarchical Decomposition}
Once we select a cluster of interest, we construct a tree by recursively clustering with HDBSCAN. For this process we use an HDBSCAN cluster epsilon of 0.1 and a minimum cluster size of 40, fixed for all experiments. The tree is constructed greedily in a depth-first search, by recursing \textit{only} on non-noise clusters. Note that because we add noise clusters back to the tree after constructing it, this can result in small disappearing regions, like in the lower levels of the succulent scene. These artifacts would better be addressed with a non-greedy tree construction, which we hope to address in future work.

To speed tree contruction, we first sub-sample the input gaussian splat with  Open3D's voxel-downsampling to reduce the resolution of points to $0.01\times$ the scale being queried, for example an affinity of 0.1 scale downsamples to .001 voxel resolution. After tree construction, the resulting tree is pruned to remove chains of nodes with one child and one parent.

\subsection{Treatment of Clustering Noise}
One challenge to overcome is the fact that HDBSCAN can output `noise' clusters, which do not get any cluster labels. %
These can arise because of gaussians which do not align well with NeRF geometry, features which are noisy because they lie on the boundary of two groups, or noise in the trained affinity field. To handle these noise clusters, we assign labels to gaussians considered noise with the label of the nearest \textit{physical} clusters computed in the Euclidean space, as opposed to the feature space. We find this produces more cohesive results than soft clustering within the feature space itself. During global clustering (Figs.~\ref{fig:scene-wide},~\ref{fig:results}) these noise clusters are assigned to custers across the entire scene, and during tree decomposition (Fig.~\ref{fig:louvre_tree}) these noise clusters are locally assigned from the clusters available at each node only.

\subsection{3D Completeness Experiment}

\subsubsection{Ground Truth Annotation}
We annotate ground truth segmentation masks on a randomly selected novel view using the online tool `Make Sense' \cite{make-sense}, employing a polygon shape for the annotation. In  Fig.~\ref{fig: label_ground_truth}, we present the visualization on our state during the data annotation process.

The annotation process begins with the assignment of a specific label point to each target object within a given view. Note that the selection of the view is randomized, involving zooming in, zooming out, or changing the angle to enhance the evaluation of view consistency effectively. These label points serve as the basis for the subsequent mask annotation, which are made at a varying level of granularity. As a case in  Fig.~\ref{fig:bouquet}, in bouquet scene, considering the click points from different angles, we annotate the masks at different hierarchical levels: the petal of the flower (fine level), the individual flower (medium level) and the whole bouquet (coarse level). For ground truth masks in other scenes, we follow similar rules, building a mask hierarchy based on the semantic meaning, ranging from fine part of the object to coarse whole object. However, note that the number of mask levels may vary depending on the complexity and the nature semantics in the scene. For example, the bear's arm in the teatime scene,  Fig.~\ref{fig:teatime}, is only annotated with two levels of hierarchy: the left hand and the whole bear. %

\subsubsection{Complete Visualizations}
\begin{figure*}
    \includegraphics[width=\textwidth]{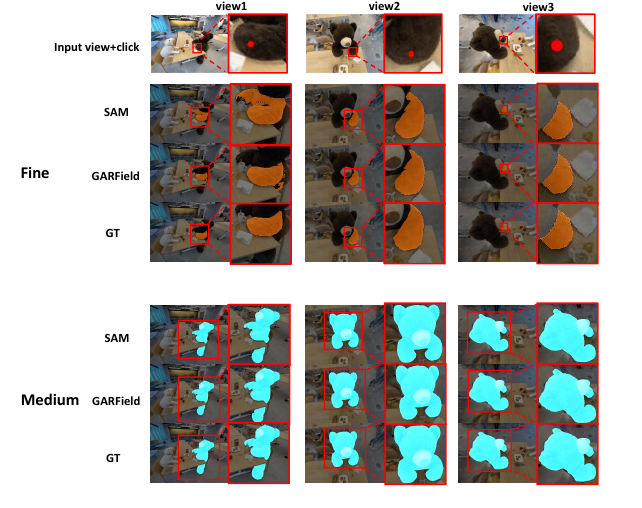}
    \caption{\textbf{View Consistency Experiment-Teatime}: We constructed two hierarchies, which are fine and medium. These correspond to the semantic meanings of the bear's left hand and the whole bear, respectively.}
    \label{fig:teatime}
\end{figure*}

\begin{figure*}
    \includegraphics[width=\textwidth]{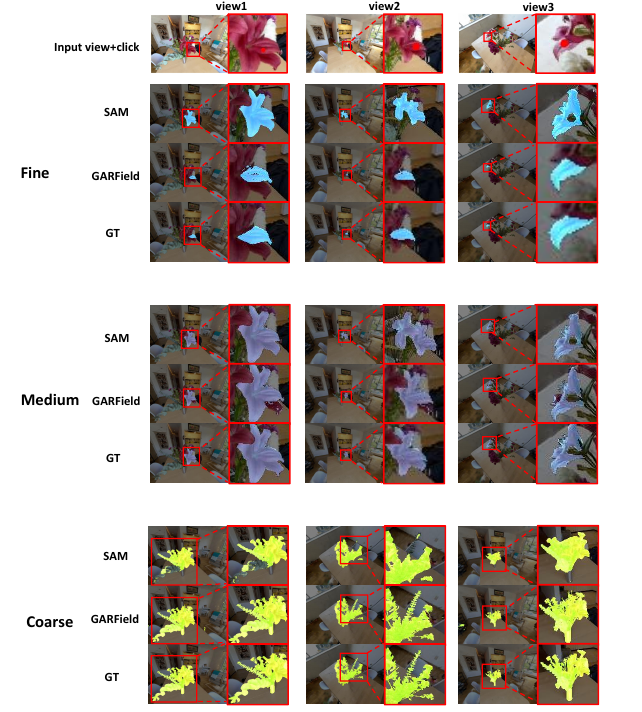}
    \caption{\textbf{View Consistency Experiment-Bouquet}: We constructed three hierarchies, which are fine medium and coarse. These correspond to the semantic meanings of the petal of the flower, the individual flower and the whole bouquet, respectively.}
    \label{fig:bouquet}
\end{figure*}

\begin{figure*}
    \includegraphics[width=\textwidth]{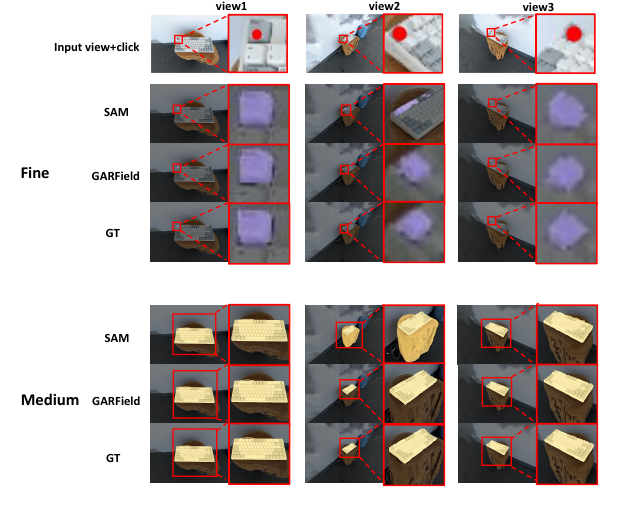}
    \caption{\textbf{View Consistency Experiment-Keyboard}: We constructed two hierarchies, which are fine and medium. These correspond to the semantic meanings of single key and the whole keyboard, respectively.}
    \label{fig:keyboard}
\end{figure*}

\begin{figure*}
    \includegraphics[width=\textwidth]{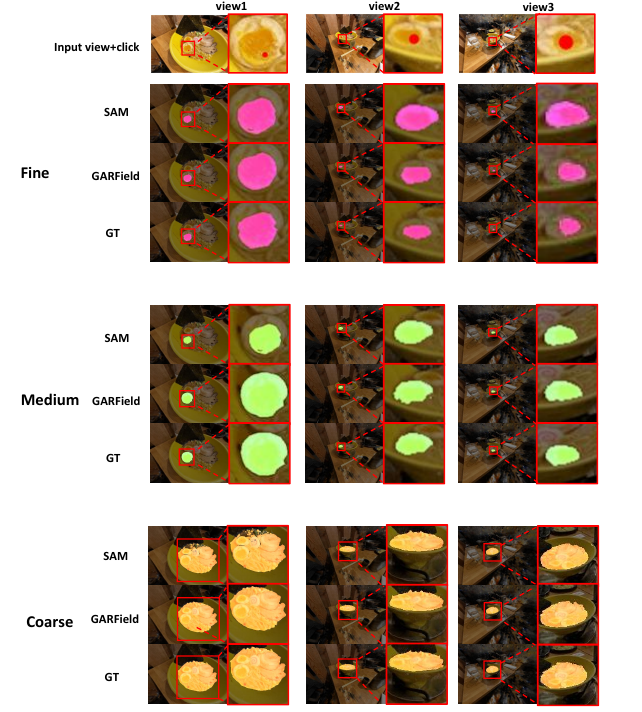}
    \caption{\textbf{View Consistency Experiment-Ramen}: We constructed three hierarchies, which are fine, medium and coarse. These correspond to the semantic meanings of egg yolk , one single egg and the whole soup area, respectively.}
    \label{fig:ramen}
\end{figure*}

\begin{figure*}
    \includegraphics[width=\textwidth]{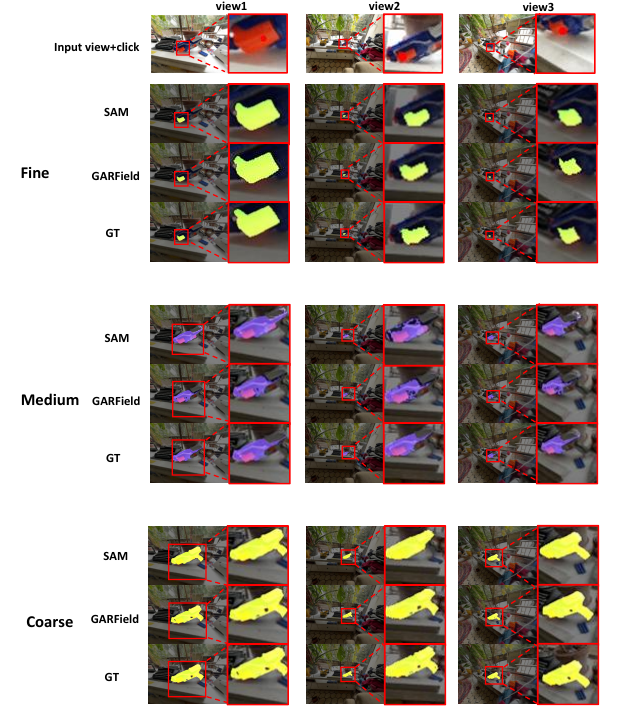}
    \caption{\textbf{View Consistency Experiment-Living room}: We constructed two hierarchies, which are fine medium and coarse. These correspond to the semantic meanings of the small orange part of the nerf gun, medium blue part of the nerf gun and the whole nerf gun, respectively.}
    \label{fig:living_room}
\end{figure*}

A comprehensive presentation of the evaluation results regarding to the view consistency of \algabbr{} is shown in  Figs.~\ref{fig:teatime},~\ref{fig:bouquet},~\ref{fig:keyboard},~\ref{fig:ramen},~\ref{fig:living_room}. This includes all the scenes not shown in the main text. For each scene, we show the clicked label points for the annotated randomly selected views, ground truth masks at different hierarchical levels and the comparison of the closest masks obtained by SAM and \algabbr{}. We also provide the zoomed-in images of the results for better visualization.

\subsection{Hierarchical Grouping Recall Experiment}

\subsubsection{Ground Truth Annotation}
In this experiment, we annotate one novel view for each of the five scenes. For each novel view, we mark one or several objects which has a rich hierarchy. The ground truth masks are any parts, subparts, or the entire object of the scene that can be considered as groups by a human. %
Taking the ramen scene (Figs.~\ref{fig: label_ground_truth},~\ref{fig:recall_results}) as an example, the parts or subparts of the objects labeled include nori, egg ,egg yolk, noodles, and so on. Additionally, the complete soup and the entire ramen bowel is also annotated as a group. Unlike the experiments on 3D completeness, this experiment aims to test whether the model can extract all the reasonable masks of the objects which contain rich hierarchy. Therefore, we did not stratify the level of the annotated masks.

\subsubsection{Complete Visualization}
\begin{figure*}
    \includegraphics[width=\textwidth]{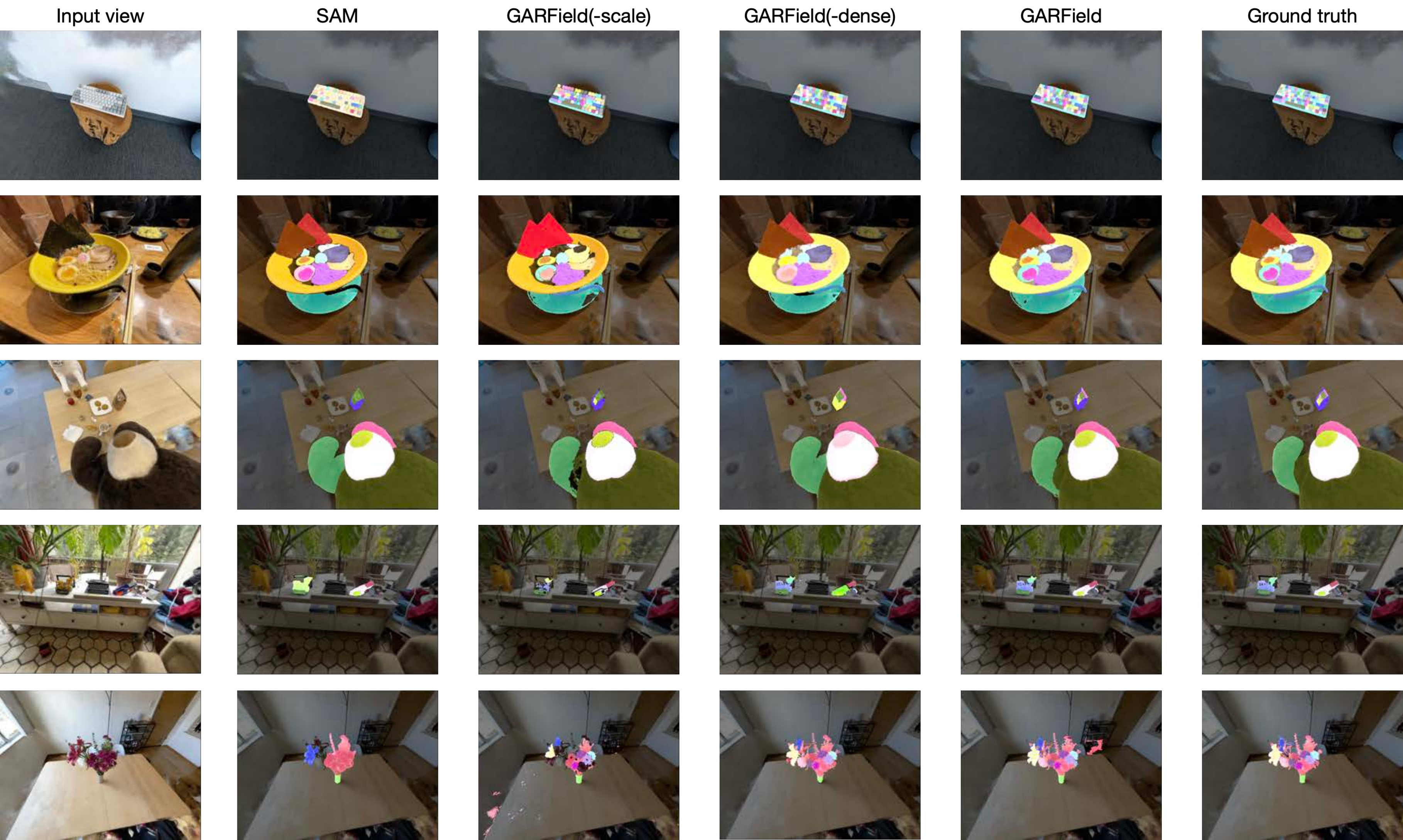}
    \caption{\textbf{Hierarchical Grouping Recall Experiments}: We concentrate on methods such as SAM and the ablation study of \algabbr{}. \algabbr{} outperforms SAM in obtaining finer, smaller masks (e.g. capturing all the tiny keys in a keyboard scene). Unlike \algabbr{} without hierarchy grouping, \algabbr{} achieves more layered grouping results (e.g. in the ramen scene, it successfully identifies the entire ramen mask through hierarchical clustering). Furthermore, compared to \algabbr{} without dense supervision, \algabbr{} provides more stable and thorough grouping outcomes (e.g. in the teatime scene, \algabbr{} more comprehensively identifies the small labels on the cookie bag).}
    \label{fig:recall_results}
\end{figure*}

In  Fig.~\ref{fig:recall_results}, We show the ground truth masks as well as all the methods masks at the finest masks. Note that all the ground truth masks are arranged in descending order of size. In our experiment, we systematically recover all the masks that corresponds to the annotated ground truth through different method. For each distinct method employed, which are SAM, \algabbr{} without scale condition, \algabbr{} without dense supervision, we sequentially showcase the masks that get the highest IOU score of the correspondence to the ground truth masks. We will release all the ground truth annotations for all experiments. 


\end{document}